\documentclass{article}

\usepackage[main, final]{neurips_2026}
 
\usepackage[utf8]{inputenc} %
\usepackage[T1]{fontenc}    %
\usepackage{hyperref}       %
\usepackage{url}            %
\usepackage{booktabs}       %
\usepackage{amsfonts}       %
\usepackage{nicefrac}       %
\usepackage{microtype}      %
\usepackage{xcolor}         %
\usepackage{graphicx}
\usepackage{subcaption}
\usepackage{float}
\usepackage{wrapfig}
\usepackage{tabularray}
\usepackage{rotating}
\usepackage{multirow} 
\usepackage{tabularx}
\usepackage{pdflscape}
\usepackage{ulem}
\usepackage[table]{xcolor}
\usepackage{adjustbox}  %
\usepackage{geometry}   %
\usepackage{enumitem}
\usepackage{bm}
\usepackage[table]{xcolor}    
\usepackage{xcolor}\usepackage{colortbl}
\usepackage{siunitx}   %
\usepackage{amsmath}
\usepackage{dsfont}
\usepackage{tikz}        %
\usepackage{xcolor}      %
\usepackage{fancyvrb}    %

\title{DASS: A Solver-Adaptable Dynamic Auxiliary Search Strategy for Symbolic Regression}

\author{Yuxiao Hu$^{1,2}$\thanks{Equal contribution.}, Qian Li$^{2,3}$\footnotemark[1], Ding Wang$^{2,3}$, Yekun Zheng$^{3}$, Jinyue Yan$^{2}$, Yuntian Chen$^{2}$\thanks{Corresponding author} \\
$^{1}$The Hong Kong Polytechnic University, Hong Kong, China\\
$^{2}$Eastern Institute of Technology, Ningbo, China\\
$^3$Shanghai Jiao Tong University, Shanghai, China\\
\texttt{yuxiao.hu@connect.polyu.hk},\texttt{qianl01205@sjtu.edu.cn}\\
}

\begin{document}

\maketitle

\begin{abstract}
Symbolic regression (SR) aims to discover interpretable mathematical expressions from data and plays a key role in scientific discovery. However, existing methods face a common bottleneck: the enormous search space and severe combinatorial explosion make it difficult to achieve a favorable trade-off between accuracy and algorithmic efficiency. We argue that, in addition to designing increasingly complex solver-specific SR strategies, an equally promising direction is to equip them with a shareable auxiliary strategy that can provide reliable search guidance. To this end, we propose DASS\footnote{\url{https://github.com/SR-Store/NeurIPS2026-DASS}}, a solver-adaptable \textbf{D}ynamic \textbf{A}uxiliary \textbf{S}earch \textbf{S}trategy that improves SR by constructing and refining a high-potential auxiliary search subspace. DASS represents the auxiliary search space as a functional subspace spanned by a set of high-potential basis function terms. It estimates term utility through multi-environment evaluation, filters unstable guidance, and updates term potential scores with a Gibbs posterior, thereby calibrating the subspace to provide more reliable guidance for downstream solvers. The high-quality solutions produced by the solver are used to update the set of basis terms. This forms a virtuous closed-loop optimization process: the subspace guides the downstream solver, and the solver feedback reshapes the subspace for subsequent exploration. DASS can be easily integrated with various types of SR solvers. Extensive experiments on LLM-SRBench show that DASS significantly improves the accuracy of original solvers with a manageable increase in runtime.
\end{abstract}

\section{Introduction}
\label{sec:intro}
Symbolic Regression (SR) is a fundamental problem in scientific discovery and interpretable modeling~\citep{chen2022symbolic,xu2023retention,meidani2023snip,zhang2025rag,li2023discrete,e2e,llmsr,li2024focus,xia2025sr}. Unlike conventional regression, which assumes a predefined functional form, SR seeks to discover the underlying mathematical expression directly from data. Given $N$ observed samples $\{{\bm x}_i,y_i\}_{i=1}^{N}$, SR aims to find a function $f:{\bm x}\mapsto y$ that explains the observed patterns. This makes SR especially important in science and engineering, where interpretability is often as crucial as predictive accuracy~\citep{yu2025discover,xu2025exploring,li2023adaptive,hu2024domain,guo2025sr,hu2025context,xu2025exploring,xu2024worth}.

\begin{figure}[htbp]
    \centering
    \includegraphics[width=1\textwidth]{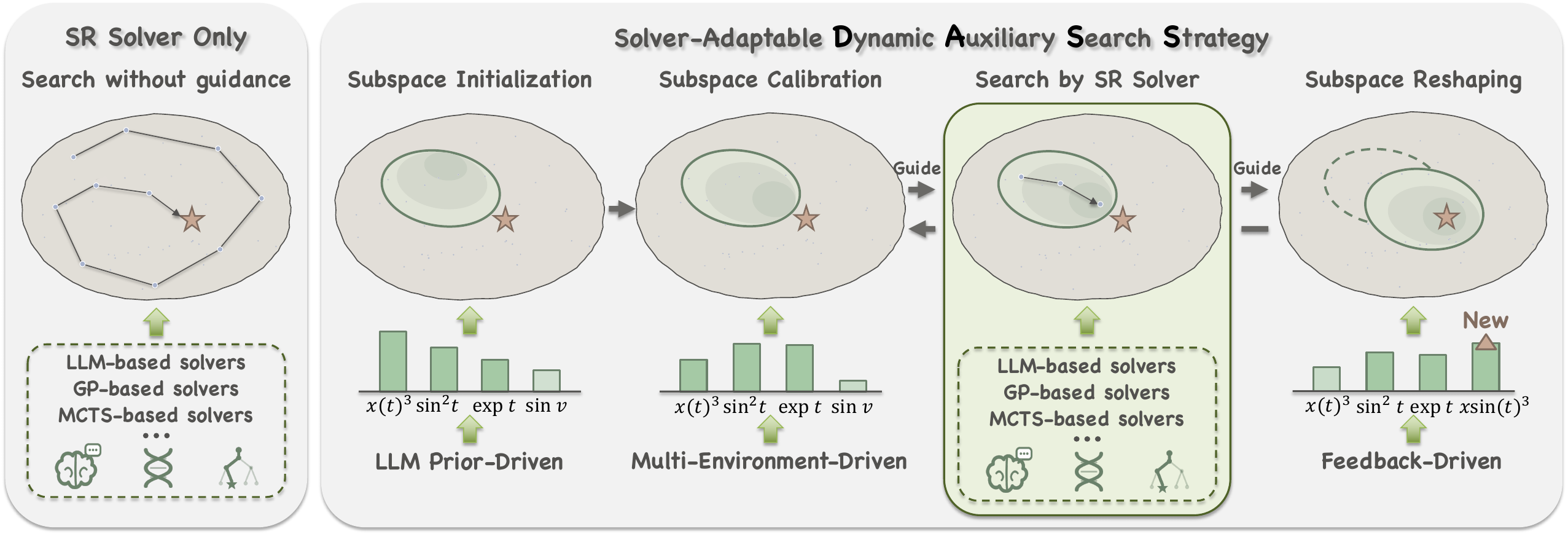}
    \caption{Core idea of DASS. Instead of letting an SR solver search the full expression space without guidance, DASS provides a solver-adaptable auxiliary search subspace spanned by high-potential basis terms. The subspace is initialized from LLM prior knowledge, calibrated by multi-environment evaluation, and reshaped using feedback from high-quality solver results. In this way, DASS guides diverse SR solvers toward promising regions while continuously refining the guided subspace.}
    \label{fig:intro}
    \vspace{-5pt}
\end{figure}

In recent years, many methods have advanced SR. However, they share a common bottleneck: the expression space is enormous and highly compositional, while effective search signals are extremely sparse~\citep{li2026gensr}. End-to-end methods~\citep{e2e,nsr,transr} regard the SR problem as a direct prediction problem from data to expressions, allowing efficient inference. Yet, when faced with such a vast combinatorial space, this one-shot prediction paradigm often struggles to ensure expression accuracy and structural reliability, as it lacks explicit structural exploration and iterative refinement. In contrast, search-based methods can often produce higher-quality results. However, whether based on Genetic Programming~\citep{pysr,gplearn,zhang2025rag}, Reinforcement Learning~\citep{dsr}, Monte Carlo Tree Search~\citep{spl}, or Large Language Models~\citep{llmsr,du2024large}, these solvers fundamentally rely on repeated trial-and-error and backtracking in a massive combinatorial space, and thus often suffer from slow search, high randomness, and low exploration efficiency. 

We argue that, beyond designing increasingly complex search strategies for individual solvers, an equally important direction is to develop a shareable auxiliary strategy that can guide the search across different solvers. Such a subspace should satisfy three key requirements: (1) \textbf{reliable}, meaning that it highlights high-potential regions that remain worth exploring across different stages and conditions rather than reflecting unstable biases induced by incidental patterns; (2) \textbf{dynamic}, meaning that it can be continuously updated by search feedback rather than providing static one-shot guidance; and (3) \textbf{solver-adaptable}, meaning that it can support various SR solvers rather than being tied to a specific SR method. As illustrated in Fig.~\ref{fig:intro}, 
with such a subspace, SR is no longer unguided trial-and-error over the full expression space; 
instead, search is steered toward high-probability regions, and optimization is lifted from improving isolated candidate expressions to continuously reshaping that region.

The key point is how to represent such an auxiliary subspace. Complete equations are too costly and inflexible as basic units, while atomic operators or symbols are too fine-grained to carry meaningful structural semantics. In this paper, we model the auxiliary subspace as an expression subspace spanned by a set of high-potential basis function terms. This term-level representation turns optimizing auxiliary subspace into the continuous updating of basis terms, enabling efficient representation and reshaping of high-potential search regions at a much lower cost.

Building on this insight, we propose DASS, a solver-adaptable \textbf{D}ynamic \textbf{A}uxiliary \textbf{S}earch \textbf{S}trategy. DASS dynamically maintains and refines a reliable, shareable auxiliary search subspace that steers the solver toward more promising regions. Specifically, DASS initializes the basis terms and their potential scores from prior knowledge, then samples expressions from the current subspace. To identify reliable search guidance, DASS evaluates these sampled expressions across multiple environments and estimates the utility of each basis term for effective search, while stability filtering suppresses terms that are only effective at particular stages or under particular environments. Based on these utility and stability signals, DASS updates term potential scores via a Gibbs posterior, thereby calibrating the subspace toward high-potential regions for prioritized exploration. 
High-quality solutions returned by the solver are then used to update the basis term set and its score distribution for subsequent rounds. This forms a virtuous closed loop: the subspace guides search, and search feedbacks in turn reshape the subspace. As a solver-adaptable front-end module, DASS can be naturally integrated with a wide range of SR solvers. 
Extensive experiments on LLM-SRBench show that, under the same search budget, DASS nearly doubles average accuracy, with a \textbf{97.84\% relative gain} and only a \textbf{24.14\% runtime increase}. Conversely, to achieve comparable or better accuracy, DASS-augmented solvers can be \textbf{1.5$\times$ to 8$\times$ faster} than their original versions.

Our contributions come in three areas: (1) We introduce a novel direction for SR: developing reliable, updatable, and shareable auxiliary search guidance for diverse solvers rather than focusing solely on solver-specific design. (2) We propose DASS, a solver-adaptable framework that models auxiliary subspace as a term-spanned search space and serves as a front-end module for different SR solvers. (3) We develop a closed-loop optimization mechanism that turns sparse solver feedback into stable term-level guidance through multi-environment evaluation, stability-aware filtering, Gibbs posterior updating, and feedback-driven subspace refinement.

\section{Related Work}
\vspace{-1em}
Most classic SR methods rely on search and optimization algorithms, such as genetic programming (GP) and its variants~\citep{pysr,gplearn,xu2020dlga,virgolin2021improving}. These methods evolve candidate equations iteratively through mutation, crossover, and selection. In addition, SPL~\citep{spl}, DSR~\citep{dsr}, and other methods utilize methods such as MCTS and RL to optimize search. However, such search-based methods require repeated exploration and backtracking in a vast discrete combinatorial space, often leading to slow search.

To improve inference efficiency, another line of work tries to directly learn the mapping from numerical observations to symbolic expressions, giving rise to end-to-end neural SR~\citep{nsr,e2e,transr}. For example, E2ESR~\citep{e2e} directly predicts complete expressions together with their constants. By shifting most of the search cost to pretraining, these methods can generate expressions rapidly during inference. However, such one-shot prediction paradigm lacks explicit exploration and correction, making it difficult to ensure reliable structures and accurate predictions.

To balance accuracy and search efficiency, several methods combine Transformer-based pre-trained models with optimization and search algorithms, aiming to leverage both learned priors and search feedback~\citep{meidani2023snip,shojaee2023transformer}. GenSR~\citep{li2026gensr} constructs a smooth latent space and uses CMA-ES algorithm to search expression. Although these methods are not purely end-to-end prediction, they still lack explicit structural guidance at the search stage.

Recently, with the development of LLMs, many works try to exploit their knowledge and program-generation capabilities for SR~\citep{xia2025sr,zhang2025gp,hu2026beyond,li2026learning}. LASR~\citep{lasr} abstracts and evolves concepts from high-quality expressions using LLMs, constructing a concept library to guide search. LLM-SR~\citep{llmsr} represents equations as programs and uses LLMs to generate equation skeletons. Actually, these methods replace traditional search processes such as crossover and mutation with language-model-driven generation. Coupled with the inherent hallucination issues of LLMs, they still require repeated exploration in an extremely large expression space and involve a large number of iterative LLM calls.

Overall, existing SR methods have made significant progress in searcher design, pretrained modeling, and the use of LLMs capabilities. While they exhibit different strengths, they share a fundamental bottleneck: the expression space is highly combinatorial and extremely large, whereas effective search feedback is sparse. Thus, we argue that instead of only improving a specific type of solver, it is more desirable to provide different solvers with a reliable, dynamic, and shareable auxiliary search strategy. In this way, search is no longer an unguided trial-and-error process over the full space, but is transformed into a continual process of guiding and reshaping high-potential regions. Thus, different solvers can achieve higher efficiency while retaining their advantages.

\section{DASS: A Solver-Adaptable Dynamic Auxiliary Search Strategy}
\label{sec:method}

\subsection{Problem Formulation}
\label{3method:problem}

Given $N$ observations ${\mathcal{\bm D}}=\{({\bm x}_i,y_i)\}_{i=1}^N$, SR aims to find an expression $f\in\mathcal{F}$ such that $f({\bm x}_i)\approx y_i$, where $\mathcal{F}$ denotes the full symbolic expression space. Since $\mathcal{F}$ is highly discrete and combinatorial, many SR solvers suffer from slow and inefficient search. To address this, we introduce a shareable auxiliary search subspace $\mathcal{S}\subset\mathcal{F}$, which captures high-potential regions of $\mathcal{F}$ and provides effective search guidance for downstream solvers.

\begin{figure}[htbp]
    \centering
    \includegraphics[width=1\textwidth]{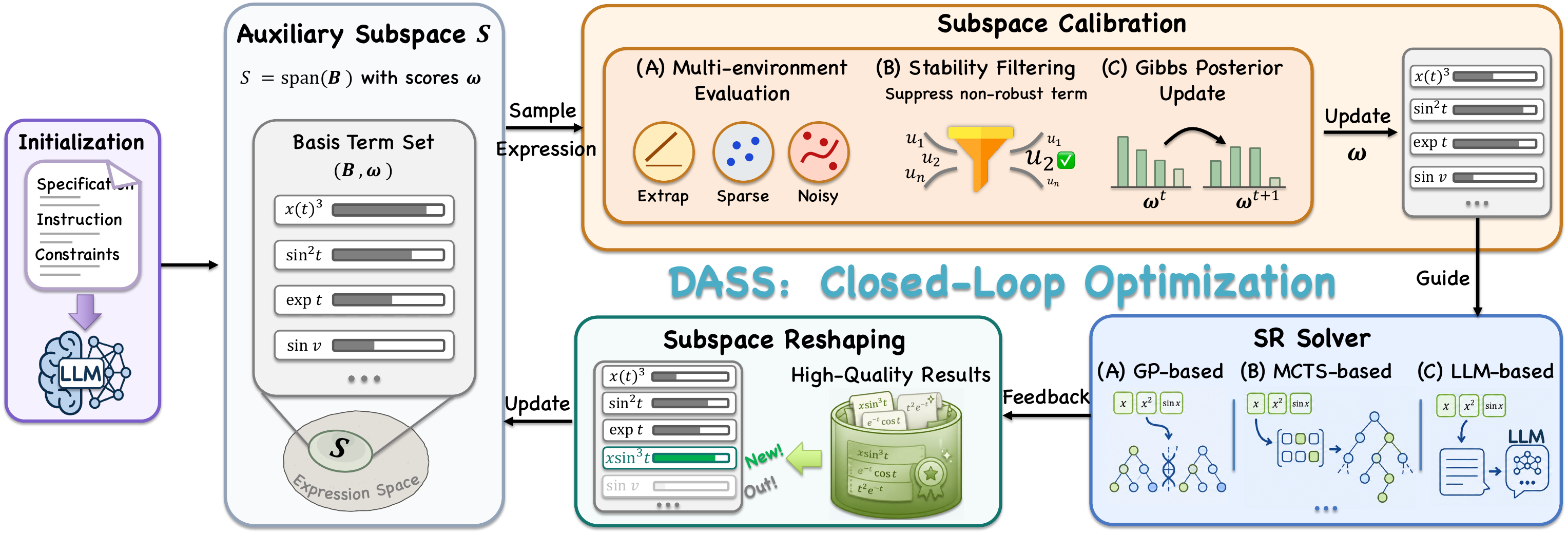}
    \caption{Framework of DASS. DASS initializes an auxiliary search subspace from prior knowledge, calibrates it through multi-environment evaluation, stability filtering, and Gibbs posterior updating, and uses the resulting high-potential terms to guide downstream SR solvers. High-quality solver outputs are then fed back to reshape the subspace, yielding a closed-loop optimization process. In addition to the three SR solvers shown in the figure, we include two additional solver types in the Appendix~\ref{Ext}: latent-space-based and policy-based solvers.}
    \label{fig:framework}
    \vspace{-5pt}
\end{figure}

Directly representing $\mathcal{S}$ with numerous complete expressions is costly and inflexible, and would again incur combinatorial explosion during subspace updates. Instead, we represent $\mathcal{S}$ using a set of basis terms $\mathcal{B}=\{s_1,\cdots, s_n\}$, where each $s_i$ is a function term, i.e., a subtree of the expression binary tree. The auxiliary subspace induced by $\mathcal{B}$ is defined in Eq.~\ref{eq}. Each basis term $s_i$ is further associated with a potential score $\omega_i$, which measures its usefulness for guiding search. Thus, $(\mathcal{B},{\bm\omega})$ provides a compact parameterization of $\mathcal{S}$ and its reliability along each basis direction. 
\begin{equation}\label{eq}
\mathcal{S}={\rm span}(\mathcal{B})=\{f\in\mathcal{F}|f \text{ can be composed from terms in }\mathcal{B}\}.
\end{equation}

DASS aims to continuously optimize the guidance induced by $\mathcal{S}$ by updating both $\mathcal{B}$ and $\bm \omega$. It consists of two coupled procedures: (1) \textbf{Subspace Calibration}, which updates $\bm \omega$ with fixed subspace structure, i.e., $\mathcal{B}$ fixed, to identify more promising directions within the subspace and provide reliable guidance to the downstream solver; and (2) \textbf{Subspace Reshaping}, which updates $\mathcal{B}$ using high-quality solutions generated by the downstream solver. These two procedures progressively improve both the auxiliary subspace and its reliability as search guidance.

\subsection{Prior-Driven Subspace Initialization}
\label{3method:InitSubspace}
At the initial stage, DASS leverages the domain-specific priors of LLMs to initialize the auxiliary subspace. Specifically, we prompt the LLM to generate commonly used function terms in the target scientific domain. The prompt includes three components: (1) role specification, which places the model in the context of classical scientific literature in the target domain; (2) task instruction, which provides the task background, variables' description, and structural constraints of candidate basis terms to be generated; and (3) output-format constraints, which define the required format to ensure that the generated results can be easily parsed and aggregated. 

To estimate the initial potential scores, we repeatedly query the LLM with the same prompt under memory-free sampling, where each generation is independent of previous outputs. Terms that appear frequently across independent generations are considered more likely under the target domain background and are assigned higher potential scores. We retain the top-$n$ unique terms by frequency to form the initial basis set $\mathcal{B}^{(0)}$, and define the initial potential score of term $s_i$ as its normalized generation frequency: $\omega^{(0)}_i=\frac{c(s_i)}{\sum^n_{j=1}c(s_j)}$, where $c(s_i)$ denotes the generation frequency of term $s_i$. The full prompt and details are provided in Appendix~\ref{app:prompt}.

\subsection{Multi-Environment-Driven Subspace Calibration}
\label{3method:Calibration}

If we regard basis terms as directions of the subspace, subspace calibration aims to estimate which directions have a higher search potential and whether such potential remains stable across different environments. We update the potential score of each term in three steps: (1) multi-environment evaluation, i.e., the utilities of basis terms are evaluated under diverse environments; (2) stability filtering, i.e., suppressing terms with unstable utility across different environments; and (3) Gibbs posterior update, i.e., we update the potential scores via Gibbs posterior. Terms with high potential scores are used to guide the downstream solver.

At $t$-th round of DASS, let $\mathcal{B}^{(t)}$ and $\bm\omega^{(t)}$ denote the current basis set and its potential scores, respectively. Since the current auxiliary subspace is induced by the terms in $\mathcal{B}^{(t)}$, we can sample $m$ expressions from the auxiliary subspace by randomly combining basis terms under a maximum-length constraint: ${\mathcal{Q}}^{(t)}=\{f_1,\cdots,f_m |f_i\sim{\rm span}(\mathcal{B}^{(t)})\}$.

These sampled expressions are then evaluated to estimate the potential scores of the terms in $\mathcal{B}^{(t)}$, thereby obtaining the score distribution over different directions of the subspace.

\textbf{Multi-Environment Evaluation.\quad}We evaluate expressions in ${\mathcal{Q}}^{(t)}$ under three environments $\mathcal{E}=\{\rm extrap, sparse, noisy\}$. Extrapolation evaluation uses one half of the data for coefficient fitting and the other half for evaluation. Sparsity evaluation fits coefficients on a sparse subset and evaluates on the remaining data. Noise evaluation adds noise before fitting and evaluation. To reduce computational cost, DASS samples and evaluates only one environment at each round.

Let $e^{(t)}\in\mathcal{E}$ be the environment sampled at round $t$ of DASS. For each sampled expression $f\in\mathcal{Q}^{(t)}$, we define its fitness as ${\rm Fitness}^{(t)}(f)=\frac{1}{1+L^{(t)}(f)}$, where $L^{(t)}(f)$ is the prediction error under $e^{(t)}$. Thus, lower error corresponds to higher fitness. We then select the top-$k\%$ expressions as the elite set $\hat{\mathcal{Q}}^{(t)}$, which serves as positive evidence for term utility.

For a basis term $s\in\mathcal{B}^{(t)}$, let $\hat{\mathcal{Q}}^{(t)}(s)\subseteq\hat{\mathcal{Q}}^{(t)}$ denote the elite expressions containing term $s$. We define the newly obtained utility of $s$ at round $t$ under environment $e^{(t)}$ as 
\begin{equation}\label{eq1}
u^{(t)}_{\rm obt.}(s)=
\begin{cases}
\dfrac{1}{|\hat{\mathcal{Q}}^{(t)}(s)|}\displaystyle\sum_{f\in \hat{\mathcal{Q}}^{(t)}(s)} \dfrac{{\rm Fitness}^{(t)}(f)}{|f|}, & |\hat{\mathcal{Q}}^{(t)}(s)|>0,\\[1em]
0, & |\hat{\mathcal{Q}^{(t)}}(s)|=0,
\end{cases}
\end{equation}
where $|f|$ is the number of terms in $f$. Eq.~\ref{eq1} rewards a term when it appears repeatedly in high-fitness expressions, while normalizing by expression length $|f|$ to avoid over-crediting terms in overly long expressions. Terms that do not appear in any elite expression receive zero utility in the current round, meaning that they are not positively supported by current search evidence.

Since $u^{(t)}_{\rm obt.}(s)$ reflects the usefulness of $s$ only in the currently sampled environment, DASS maintains a cross-environment utility state for each term: $[u^{(t)}_{\rm extrap}(s),u^{(t)}_{\rm sparse}(s),u^{(t)}_{\rm noisy}(s)]$, which stores the utility of $s$ across the three environments at round $t$. Considering the diminishing influence of earlier rounds, DASS updates the utility state using an exponential moving average (EMA): 
\begin{equation}\label{ema}
u_{e}^{(t)}(s)=
\begin{cases}
(1-\eta)\,u_{e}^{(t-1)}(s)+\eta\,u^{(t)}_{\rm obt.}(s), & e=e^{(t)},\\[0.8em]
u_{e}^{(t-1)}(s), & e\ne e^{(t)},
\end{cases}
\end{equation}
where $\eta$ is the update rate. In this way, DASS tracks each term’s utility across the three environments, respectively, without requiring all three environments to be evaluated at every round.

We then summarize the environment-wise utility state into a comprehensive utility:

\begin{equation}\label{eq2}
u^{(t)}_{\rm com.}(s)=\frac{1}{|\mathcal{E}|}\sum_{e\in\mathcal{E}}u_e^{(t)}(s).
\end{equation}
This provides an overall estimate of how useful term $s$ is across environments in DASS round $t$.

\textbf{Stability Filtering.\quad}
Utility alone is insufficient for reliable guidance. A term may achieve high utility in one environment but perform poorly in others, in which case promoting it would distort the auxiliary subspace toward unstable directions. DASS therefore introduces a stability filter. We measure the cross-environment stability of a term by: 
$$
\gamma^{(t)}(s)=\frac{\min_{e\in\mathcal{E}}u_e^{(t)}(s)}{u^{(t)}_{\rm com.}(s)+\epsilon},
$$
where $\epsilon$ is a small constant for numerical stability. This formulation indicates that $\gamma^{(t)}(s)$ approaches 1 when a basis term exhibits stable utility across different environments, and approaches 0 otherwise. We then define the final utility under the stability constraint as: 
$$
u^{(t)}(s)=u^{(t)}_{\rm com.}(s)\cdot\gamma^{(t)}(s).
$$
This stability-constrained utility ensures that only terms with both high average utility and cross-environment consistency are strongly reinforced.

\textbf{Gibbs Posterior Update.\quad} %
After obtaining the final utility $u^{(t)}(s)$  at round $t$, DASS updates the potential scores of basis terms through a Gibbs posterior:
$$
\omega^{(t+1)}(s)\propto\omega^{(t)}(s)\cdot{\rm exp}(\beta u^{(t)}(s)),
$$
followed by normalization over all basis terms. The purpose of this update is to reweight the current potential-score distribution using the utility signal obtained at round $t$, so that potential-score distribution is progressively shifted toward more useful and stable terms. Here, $\beta$ controls the strength of the current search evidence in reshaping the distribution. Terms not supported by any elite expression receive zero observed utility in the current round, i.e. $u^{(t)}_{\rm obt.}(s)=0$, which decreases $u^{(t)}_{\rm com.}(s)$ and $u^{(t)}(s)$. 
Thus, such terms may remain temporarily, but if they continue to receive no utility in subsequent rounds, feedback-driven subspace reshaping removes them and steers the auxiliary subspace toward high-utility, cross-environment-stable directions.

\subsection{Solver-Adaptable Search Guidance}
\label{3method:DownSolver}

After calibration, DASS obtains a set of basis terms with updated potential scores, which define the current auxiliary search subspace. To guide downstream symbolic regression solvers, DASS samples high-potential terms according to these scores and converts them into solver-specific search guidance direction. For LLM-based solvers, these terms are injected into prompts as structural hints; for evolutionary solvers, they bias mutation and crossover toward promising symbolic components; and for MCTS-based solvers, they are incorporated into the grammar library to guide tree expansion. In all cases, DASS acts as a front-end module rather than a replacement for the solver itself. Additional implementation details are deferred to the Appendix~\ref{app:dass-integration}.

\subsection{Feedback-Driven Subspace Reshaping}
DASS is invoked periodically rather than activated at every solver step, which both limits the overhead of subspace calibration and allows the downstream solver sufficient time to adapt to the current search signal. Between two DASS rounds $t$ and $t+1$, the solver runs with its native search mechanism and produces candidate expressions independently. These solver-generated high-quality expressions are then used as structural feedback to reshape the auxiliary subspace for the next DASS round.

Specifically, let $\hat{\mathcal{P}}^{(t)}$ denote the set of high-quality expressions generated during the DASS-free phase after the $t$-th DASS stage. We retain those expressions whose fitness is no lower than the minimum fitness in the elite set $\hat{\mathcal{Q}}^{(t)}$, i.e., $\hat{\mathcal{P}}^{(t)}=\left\{f \mid \mathrm{Fitness}(f)\geq \min\big(\mathrm{Fitness}(\hat{\mathcal{Q}}^{(t)})\big)\right\}$. These expressions are treated as positive evidence for discovering useful new terms that are not yet included in the current basis term set $\mathcal{B}^{(t)}$.

For a new term $s\notin\mathcal{B}^{(t)}$, let $\hat{\mathcal{P}}^{(t)}(s)=\{f\in\hat{\mathcal{P}}^{(t)} \mid f \text{ contains term } s\}$ denote the subset of retained expressions containing $s$. Similarly, we define the utility of this new term from solver feedback as

$$
u(s)=
\begin{cases}
\dfrac{1}{|\hat{\mathcal{P}}^{(t)}(s)|}\displaystyle\sum_{f\in \hat{\mathcal{P}}^{(t)}(s)} \dfrac{{\rm Fitness}(f)}{|f|}, & |\hat{\mathcal{P}}^{(t)}(s)|>0,\\[1em]
0, & |\hat{\mathcal{P}^{(t)}}(s)|=0.
\end{cases}
$$

DASS then reshapes the auxiliary subspace by replacing low-utility basis terms in $\mathcal{B}^{(t)}$ with high-utility new terms, while keeping the set size $|\mathcal{B}|$ unchanged. The initial potential score of each inserted term is set according to $\omega_{\mathrm{new}}^{(t+1)}(s)\propto \omega_0 \exp\big(\beta u^{(t)}_{\mathrm{new}}(s)\big)$, where $\omega_0$ is a fixed hyperparameter controlling the initial preference for newly introduced terms. Before the $(t+1)$-th DASS round, the potential scores over the updated basis term set are normalized and used as the initial score distribution of the next-stage auxiliary subspace.

\section{Experiment}
\label{sec:exp}

\subsection{Experimental Setup}
\paragraph{Dataset.}
Since our method and the main baselines involve LLMs, we evaluate DASS on the LLM-SRBench~\citep{llmsrbench}. As noted in~\cite{xia2025sr}, we also find that LSR-Transform dataset of LLM-SRBench may still suffer from memorization issues. Thus, we choose LSR-Synth dataset of LLM-SRBench to evaluate all methods. LSR-Synth dataset constructs symbolic regression problems by combining known functional terms with synthetically generated novel terms, thereby avoiding evaluation contamination from equations that may have been memorized by LLMs during large-scale pretraining. LSR-Synth contains $129$ symbolic regression tasks spanning four scientific domains: Physics ($44$), Materials Science ($25$), Chemistry ($36$), and Biology ($24$). Each task provides a training set, an in-domain (ID) test set, and an out-of-domain (OOD) test set.

\paragraph{Metrics.}
We report the average search time per problem recorded as Time (seconds), the normalized mean squared error (NMSE), and $\mathrm{Acc}_{0.01}$ following~\cite{e2e};~\cite{xia2025sr};~\cite{llmsr} to show the task-level solving accuracy. Given predictions $\hat{y}_i = f(\bm{x}_i)$ and ground-truth values $y_i$, NMSE is defined as
\[
\mathrm{NMSE}(f)
=
\frac{
\sum_{i=1}^{N_{\mathrm{test}}} (\hat{y}_i - y_i)^2
}{
\sum_{i=1}^{N_{\mathrm{test}}} (y_i - \bar{y})^2
}.
\]
For $\mathrm{Acc}_{0.01}$, we follow the accuracy-to-tolerance definition in LLM-SRBench: a formula is counted as correct if the maximum relative error over all test points does not exceed a tolerance threshold $\tau$,
\[
\mathrm{Acc}_{\tau}(f)
=
\mathds{1}
\left(
\max_{1 \le i \le N_{\mathrm{test}}}
\left|
\frac{\hat{y}_i - y_i}{y_i}
\right|
\le \tau
\right).
\]
In our experiments we set $\tau = 0.01$, corresponding to $1\%$. To prevent the $\max$ operator from being dominated by a small number of outliers, we follow \citet{nsr,e2e,xia2025sr} and discard the worst $5\%$ of test points for each task.

\paragraph{Base solvers.}
DASS is not designed as a standalone SR solver, but rather as a plug-in module that assists existing solvers. The position of DASS makes our main experiments substantially different from prior SR work. Under the same settings, comparing solver A with solver A+DASS most directly isolates the gain introduced by DASS, which motivates the paired comparisons in Tab.~\ref{tab:main-results}. In contrast, comparing solver A+DASS with solver B, conflates the effects of both the base solver and DASS. Similarly, comparing solver A+DASS with solver B+DASS still includes the inherent performance gap between solvers A and B. A direct comparison between DASS and solver A is not meaningful because DASS cannot perform symbolic search independently.

In the main text, we integrate DASS into five representative SR solvers covering three search paradigms: LLM-based solvers, including LLM-SR~\citep{llmsr} and LaSR~\citep{lasr}; GP-based solvers, including PySR~\citep{pysr} and gplearn~\citep{gplearn}; and an MCTS-based solver, SPL~\citep{spl}. For the two LLM-based methods, we evaluate them and their DASS-augmented variants on both a small LLM, Qwen3.5-9B~\citep{qwen3.5}, and a larger LLM, Gemma-4-31B. For each solver, the baseline and its DASS-augmented variant use exactly the same core search hyperparameters, so that the comparison isolates the contribution of the DASS module itself. Moreover, the Appendix~\ref{Ext} provides implementation details and experimental results for integrating DASS with latent-space-based method (GenSR~\citep{li2026gensr}) and policy-based method (DSR~\citep{dsr}). These two methods do not expose an explicit interface for injecting symbolic priors or search biases. Nevertheless, DASS can still be coupled to such methods by intervening at the latent-search level or at the policy level. 

\paragraph{Search budget and DASS configuration.}
The search budgets of the solvers are set as follows: for each problem, we set the search budgets to $100$ generated candidates for LLM-SR, $20$ outer iterations for LaSR, $65$ outer iterations for PySR, $100$ generations for gplearn, and $10$ transplants for SPL. Unless otherwise specified, these values are used as the default settings. This setting is intended to evaluate whether DASS can help base solvers converge more efficiently. Under large budgets, base solvers approach saturation, creating a ceiling effect that compresses final accuracy differences and obscures DASS’s ability to reach comparable performance earlier. Fig.~\ref{fig:budget-compare} and Fig.~\ref{fig:phys_budget} to~\ref{fig:bio_budget} therefore compare base solver+DASS under a low budget with the base solver under a high budget. DASS-enhanced solvers match or even exceed the high-budget base solvers. 

DASS randomly samples $m = 200$ expressions from the current auxiliary subspace for multi-environment evaluation, and selects the top $30\%$ by fitness, as the elite set $\hat{\mathcal{Q}}^{(t)}$. The capacity of the basis term set is fixed at $n=|\mathcal{B}| = 25$. For reshaping, the initial potential score of newly introduced terms is set to $\omega_0 = 0.5$.

Each experiment is repeated with three different random seeds. Unless otherwise specified, the budget-efficiency analysis and all ablations use Qwen3.5-9B for LLM-based solvers. Additional details on the dataset, base solvers, and solver-specific hyperparameters are provided in Appendices~\ref{app:dataset},~\ref{app:base-solvers}; OOD results, per-domain budget analysis, and ablation/sensitivity studies are provided in Appendices~\ref{app:additional-results}, ~\ref{app:sensitivity}.

\subsection{Results of Different SR Solvers Integrated with DASS}
\label{4exp:mainres}

\begin{table}[htbp]
  \centering
  \caption
    {
    Main results on the four domain-specific datasets in LLM-SRBench.
    We evaluate DASS by integrating it into five representative SR solvers and report the average search time (Time), normalized mean squared error (NMSE), and $\mathrm{Acc}_{0.01}$.
    Shaded rows indicate DASS-augmented variants, and boldface denotes better value within each base solver / +DASS pair. 
    }
  \label{tab:main-results}
  \resizebox{\linewidth}{!}{\begingroup
\def\g{\cellcolor{gray!10}}%
\begin{tabular}{llcccccccccccc}
\toprule
\multirow{2}[4]{*}{\textbf{Category}}
  & \multirow{2}[4]{*}{\textbf{Method}}
  & \multicolumn{3}{c}{\textbf{Phys}}
  & \multicolumn{3}{c}{\textbf{MatSci}}
  & \multicolumn{3}{c}{\textbf{Chem}}
  & \multicolumn{3}{c}{\textbf{Bio}} \\
\cmidrule(lr){3-5}\cmidrule(lr){6-8}\cmidrule(lr){9-11}\cmidrule(lr){12-14}
 & & Time$\downarrow$ & NMSE$\downarrow$ & $\mathrm{Acc}_{0.01}$$\uparrow$
   & Time$\downarrow$ & NMSE$\downarrow$ & $\mathrm{Acc}_{0.01}$$\uparrow$
   & Time$\downarrow$ & NMSE$\downarrow$ & $\mathrm{Acc}_{0.01}$$\uparrow$
   & Time$\downarrow$ & NMSE$\downarrow$ & $\mathrm{Acc}_{0.01}$$\uparrow$ \\
\midrule
\multirow{10}{*}[-10pt]{\textbf{LLM-based}}
 & \multicolumn{13}{c}{Qwen3.5-9B} \\
\cmidrule(lr){2-14}
 & LLMSR                      & 197.2 & 5.32E-03 & 9.1 & 185 & 2.38E-05 & 38.7 & 177.3 & 4.96E-04 & 7.4 & 187.9 & 2.14E-03 & 6.9 \\
 & \g\textbf{LLMSR+DASS}    & \g205.1 & \g\textbf{6.04E-04} & \g\textbf{29.5} & \g257.2 & \g\textbf{2.91E-06} & \g\textbf{81.3} & \g260.4 & \g\textbf{1.15E-04} & \g\textbf{23.1} & \g284.2 & \g\textbf{4.52E-04} & \g\textbf{13.9} \\
 & LASR                       & 413.5 & 2.24E-02 & 6.8 & 502.7 & 6.76E-04 & 16.0 & 477.1 & 5.24E-03 & 13.9 & 429 & 5.64E-02 & 6.9 \\
 & \g\textbf{LASR+DASS}    & \g523.8 & \g\textbf{4.18E-03} & \g\textbf{18.9} & \g566.9 & \g\textbf{8.15E-05} & \g\textbf{41.3} & \g496.2 & \g\textbf{6.24E-04} & \g\textbf{18.5} & \g491.3 & \g\textbf{7.64E-03} & \g\textbf{12.5} \\
\cmidrule(lr){2-14}
 & \multicolumn{13}{c}{Gemma-4-31B} \\
\cmidrule(lr){2-14}
 & LLMSR                      & 191.6 & 2.19E-03 & 13.6 & 194.3 & 3.23E-05 & 45.3 & 188.7 & 1.01E-04 & 17.6 & 203 & 9.14E-04 & 11.1 \\
 & \g\textbf{LLMSR+DASS}    & \g233.4 & \g\textbf{3.42E-04} & \g\textbf{36.4} & \g275.2 & \g\textbf{1.15E-05} & \g\textbf{64.0} & \g256.2 & \g\textbf{7.11E-05} & \g\textbf{28.7} & \g298.6 & \g\textbf{2.73E-04} & \g\textbf{18.1} \\
 & LASR                       & 420.7 & 8.16E-03 & 11.4 & 491.7 & 5.18E-04 & 24.0 & 492.3 & 3.76E-03 & 17.6 & 482.8 & 1.93E-02 & 8.3 \\
 & \g\textbf{LASR+DASS}    & \g538.2 & \g\textbf{2.75E-03} & \g\textbf{23.5} & \g573.6 & \g\textbf{6.17E-05} & \g\textbf{45.3} & \g542.1 & \g\textbf{5.29E-04} & \g\textbf{23.1} & \g567.4 & \g\textbf{7.72E-03} & \g\textbf{13.9} \\
\midrule
\multirow{4}{*}{\textbf{GP-based}}
 & PySR                       & 71.7 & 4.11E-03 & 18.2 & 73.5 & 2.01E-04 & 24.0 & 81.1 & 8.93E-04 & 11.1 & 77.8 & 5.29E-03 & 20.8 \\
 & \g\textbf{PySR+DASS}    & \g92.7 & \g\textbf{1.45E-03} & \g\textbf{22.7} & \g89.3 & \g\textbf{3.80E-05} & \g\textbf{36.0} & \g97.8 & \g\textbf{4.24E-04} & \g\textbf{25.0} & \g104.6 & \g\textbf{1.48E-03} & \g\textbf{33.3} \\
 & GPlearn                    & 46.7 & 8.51E-02 & 0.0 & 52.5 & 2.36E-02 & 0.0 & 42.6 & 6.56E-02 & 0.0 & 63.4 & 4.42E-01 & 0.0 \\
 & \g\textbf{GPlearn+DASS}    & \g64.2 & \g\textbf{2.71E-02} & \g 0.0 & \g68 & \g\textbf{7.49E-03} & \g\textbf{4.0} & \g59.3 & \g\textbf{2.13E-02} & \g 0.0 & \g69.2 & \g\textbf{4.44E-02} & \g 0.0 \\
\midrule
\multirow{2}{*}{\textbf{MCTS-based}}
 & SPL                        & 593.4 & 8.41E-03 & 6.8 & 554.8 & 5.62E-04 & 20.0 & 603.6 & 9.49E-03 & 15.7 & 575.2 & 7.41E-02 & 8.3 \\
 & \g\textbf{SPL+DASS}    & \g670.9 & \g\textbf{2.82E-03} & \g\textbf{22.0} & \g646.1 & \g\textbf{7.24E-05} & \g\textbf{26.7} & \g682 & \g\textbf{4.02E-03} & \g\textbf{24.1} & \g651.8 & \g\textbf{6.74E-03} & \g\textbf{12.5} \\
\bottomrule
\end{tabular}%
\endgroup
}
  \vspace{-5pt}
\end{table}

Table~\ref{tab:main-results} compares five representative solvers and their DASS-augmented variants on LSR-Synth of LLM-SRBench. These five solvers cover three kinds of SR search paradigms, which provide interfaces that allow integrating with DASS. For SR solvers without explicit guidance-receiving interfaces, integrating DASS may require minor intervention to the original solver. In the Appendix~\ref{Ext}, we provide results and details for integrating DASS with latent-space-based and policy-based SR solvers. In this section, DASS preserves the original core mechanisms of these solvers, while substantially improving NMSE and $\mathrm{Acc}_{0.01}$ on every solver. An exception is GPlearn, where the $0\%$ baseline reflects a structural gap that guidance alone cannot bridge. On average, DASS improves NMSE and $\mathrm{Acc}_{0.01}$ by $5.27{\times}$ and $1.98{\times}$, respectively. Overall, DASS consistently improves performance across three structurally distinct search paradigms. Since DASS is an additional module for SR solvers, it inevitably introduces extra time overhead. Averaged over all solvers, the search time increases by $24.14\%$.
This overhead is acceptable given the gains, and is further controlled by DASS’s periodic activation, which also allows downstream solvers sufficient time to exploit the guidance.

\subsection{Low-Budget Base Solvers with DASS vs. High-Budget Base Solvers}
\label{4exp:budget}

\vspace{-5pt}

Table~\ref{tab:main-results} shows that DASS consistently improves the accuracy of the corresponding base solvers under a fixed search budget. In practice, however, search budget is often a primary concern~\citep{yusymbolic}: LLM calls incur non-negligible cost, and non-LLM solvers can also require long runtimes on individual problems. Therefore, a more meaningful comparison is whether a DASS-augmented solver with low-budget can match a DASS-free base solver with high-budget.

\begin{figure}[H]
  \centering
  \includegraphics[width=\linewidth]{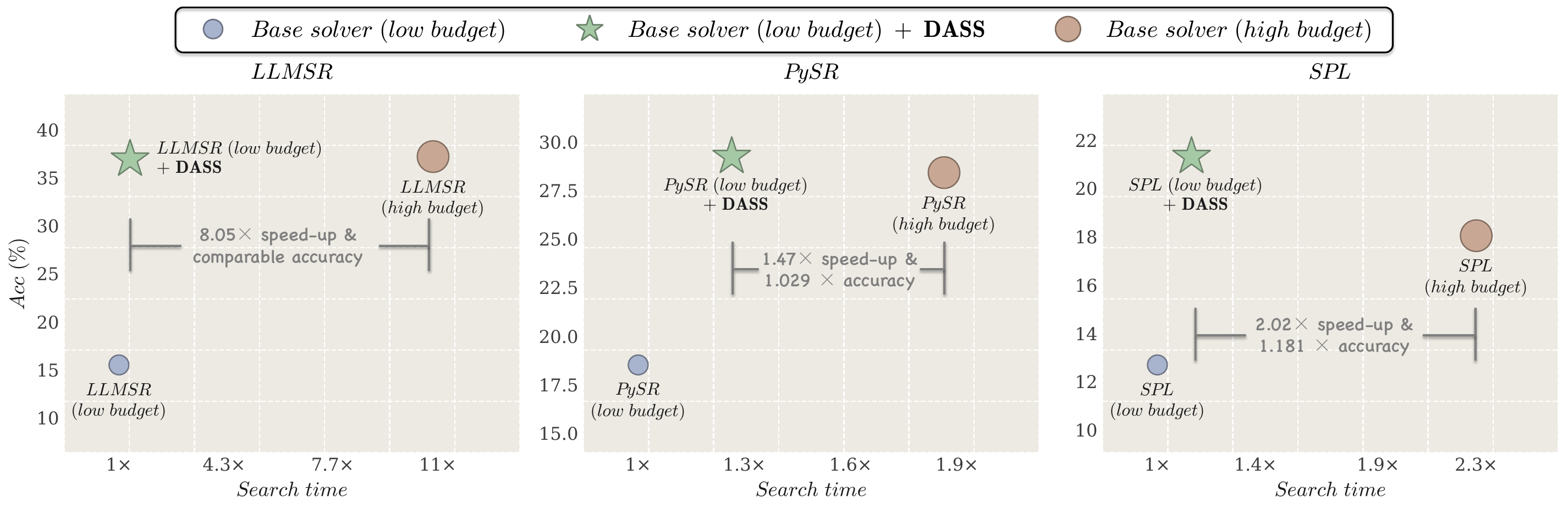}
  \caption{
    Low-budget DASS versus high-budget base solvers on LLMSR, PySR, and SPL. Each solver is evaluated under low and high search budgets with all other hyperparameters fixed; the high-budget setting increases LLMSR max LLM calls from $100$ to $1000$, PySR outer iterations from $65$ to $130$, and SPL MCTS transplants from $10$ to $20$. Markers indicate low-budget base solver (light-blue circle), low-budget DASS-augmented solver (green star), and high-budget base solver (large brown circle). Results are averaged over four domains. The $x$-axis gives relative search time normalized by the low-budget base solver ($1{\times}$), and the $y$-axis gives $\mathrm{Acc}_{0.01}$.
    }
  \vspace{-5pt}
  \label{fig:budget-compare}
\end{figure}

We answer this question using three representative solvers from distinct search paradigms: LLMSR, PySR, and SPL. Each solver uses low and high budgets, with all other hyperparameters fixed; details are in Figure~\ref{fig:budget-compare}. Averaged over the four scientific domains, Figure~\ref{fig:budget-compare} shows a consistent trend: low-budget DASS matches or exceeds the accuracy of the corresponding high-budget base solver while requiring substantially less search time. For LLMSR and PySR, DASS reaches comparable accuracy using only about $12.42\%$ and $67.87\%$ of the high-budget search time, respectively. For SPL, it uses about $49.53\%$ of the time and further improves accuracy by $18.17\%$. This result indicates that by shifting the solver away from undirected search over the full expression space and into a continuously improving high-potential subspace, DASS extracts more useful signal per unit of search budget. In these settings, simply scaling up the budget is less effective than adding DASS guidance.

\subsection{Ablation of Multi-Environment Utility Aggregation} %

\vspace{-5pt}

\begin{table}[htbp]
  \centering
  \vspace{-5pt}
  \caption{
    Ablation study on multi-environment evaluation, measured by $\mathrm{Acc}_{0.01}$.
    All other hyperparameters are kept fixed, and we vary only the choice of environments used to compute utility.
    (1) \emph{Single env:} evaluates utility using only one environment in $\mathcal{E}=\{\rm extrap, sparse, noisy\}$ .
    (2) \emph{All three envs:} the DASS setting, which alternates sampling across the three evaluation environments.
  }
  \label{tab:multi-env-ablation}
  \resizebox{\linewidth}{!}{\begin{tabular}{ll cccc cccc cccc}
\toprule
\multirow{2}[3]{*}{\textbf{Setting}} & \multirow{2}[3]{*}{\textbf{Variant}}
  & \multicolumn{4}{c}{\textbf{LLMSR}}
  & \multicolumn{4}{c}{\textbf{PySR}}
  & \multicolumn{4}{c}{\textbf{SPL}} \\
\cmidrule(lr){3-6}\cmidrule(lr){7-10}\cmidrule(lr){11-14}
 & & Phys & MatSci & Chem & Bio
   & Phys & MatSci & Chem & Bio
   & Phys & MatSci & Chem & Bio \\
\midrule
\multirow{3}{*}{\textbf{Single-env.\ utility}}
   & Extrap   & 26.5 & 74.7 & 20.4 & \textbf{9.7} & 21.2 & 32.0 & 21.3 & 30.6 & 18.2 & 25.3 & 21.3 & \textbf{12.5} \\
   & Sparse & 22.0 & 66.7 & 15.7 & 8.3          & 20.5 & 29.3 & 19.4 & 26.4 & 13.6 & 22.7 & 19.4 & 11.1 \\
   & Noisy    & 24.2 & 72.0 & 21.3 & 8.3          & 21.2 & 30.7 & 18.5 & 27.8 & 16.7 & 22.7 & 22.2 & 11.1 \\
\midrule
\multicolumn{2}{l}{\textbf{Multi-env.\ utility (DASS)}}
   & \textbf{29.5} & \textbf{81.3} & \textbf{23.1} & \textbf{13.9}
   & \textbf{22.7} & \textbf{36.0} & \textbf{25.0} & \textbf{33.3}
   & \textbf{22.0} & \textbf{26.7} & \textbf{24.1} & \textbf{12.5} \\
\bottomrule
\end{tabular}%
}
\end{table}

The four settings in Table~\ref{tab:multi-env-ablation} differ only in the environments used to evaluate the utility $u^{(t)}_{\rm com.}(s)$. The three single-environment baselines use one environment from $\mathcal{E}=\{\rm extrap, sparse, noisy\}$, whereas DASS rotates across all three following Eqs.~\ref{eq1}--\ref{eq2} in Section~\ref{3method:Calibration}; since only one environment is evaluated per round, DASS matches the single-environment baselines in per-round evaluation procedure. Against the best single-environment baseline, multi-environment evaluation improves $\mathrm{Acc}_{0.01}$ by $9.11\%$ on average, and no single-environment setting outperforms DASS. This consistent and robust advantage indicates that single-environment utility can be biased, while stability-constrained cross-environment evaluation suppresses unstable signals arising from terms that appear useful only under a specific environment, thereby further improving performance.

The three environments also differ in isolation. Extrap is the strongest single signal in most cases, possibly because extrapolation evaluation is most sensitive to the generalization ability of candidate terms. Noisy is close to Extrap, whereas Sparse is the weakest overall. Nevertheless, even the strongest single environment, Extrap, is still generally inferior to DASS. This suggests that the three environments are complementary rather than redundant: each captures a distinct stability aspect, and their aggregation provides information unavailable from any single environment.

\section{Conclusion: Limitations \& Future Work}
\label{sec:conclusion}

We introduced DASS, a solver-adaptable dynamic auxiliary search strategy for SR. DASS maintains a continuously updated auxiliary search subspace that guides the search toward high-potential regions of the expression space. 
Nevertheless, DASS has limitations. Although DASS is designed to be solver-adaptable, its integration still depends on whether the downstream solver provides an interface for receiving guidance signals. For example, genetic algorithms can incorporate guidance signals during crossover and mutation; LLM-based methods can include them in prompts; and MCTS methods explicitly maintain a grammar library for priors. For solvers without explicit prior-injection or bias-control mechanisms, additional design may be needed before DASS can be applied effectively. For example, DASS can still be coupled to latent-space-based and policy-based methods by intervening at the latent-search level or at the policy level, which are discussed in Appendix~\ref{Ext}.

More broadly, we hope this work highlights a complementary direction for symbolic regression: beyond building stronger standalone solvers, it is also valuable to study shared, reliable, and updatable search guidance across solver families. We hope based on our DASS, future work can explore more universal guidance signals, richer subspace representations, and extensions of this closed-loop paradigm to broader scientific discovery settings.

\section{Acknowledgement}
This work was supported by the National Natural Science Foundation of China (12572266), and the High Performance Computing Centers at Eastern Institute of Technology, Ningbo, and Ningbo Institute of Digital Twin.

\bibliography{ref}
\bibliographystyle{plainnat}

\newpage
\appendix

\section{Dataset Details}
\label{app:dataset}

Symbolic-regression benchmarks have evolved from physics-inspired equation discovery~\citep{udrescu2020aifeynman,lacava2021srbench} to more recent LLM-driven scientific equation discovery~\citep{llmsrbench}. Each task in LSR-Synth~\citep{llmsrbench} is defined by a synthetic ground-truth expression together with its corresponding set of input variables. The expression-generation pipeline proceeds as follows: GPT-4o~\citep{gpt4o} first generates a set of candidate synthetic terms for a given scientific question; these candidates are then sequentially filtered by a numerical solver to verify solvability, by GPT-4o to assess novelty, and finally by domain experts, who cross-validate the retained expressions against their data visualizations.

Each task contains 4,000 training samples, 500 in-distribution (ID) test samples, and 500 out-of-distribution (OOD) test samples. Gaussian noise, whose variance is scaled according to the magnitude of the target variable, is added to the outputs. The ID test samples are drawn independently from the same variable ranges as the training set. In contrast, the OOD test samples extend the range of at least one variable (typically time or a control variable) beyond the training interval, primarily via upper-end extrapolation, in order to evaluate the stability of the fitted expression under variable extrapolation. All tasks in LSR-Synth provide both ID and OOD test sets, and Appendix~\ref{app:ood} reports OOD performance separately.

\section{Base Solvers, LLM Backbones, and Compute Resources}
\label{app:base-solvers}

\subsection{Base Solver Configurations}
\label{app:base-solver-configs}

\paragraph{LLM-SR}

LLM-SR~\citep{llmsr} represents an equation as a Python program with constant placeholders: the LLM iteratively proposes program skeletons, while a numerical optimizer fits the constants for each skeleton~\citep{bfgs}. High-scoring skeletons are then added to an experience pool composed of multiple sampling islands~\citep{multi-islands, novikov2025alphaevolve}, which is used to guide subsequent proposals. In our configuration, LLM-SR generates at most 100 candidate programs per task, a $10{\times}$ reduction from the 1000 used in the original paper; this budget is kept identical between the base solver and its DASS-augmented variant. As shown in Section~\ref{4exp:budget}, the base solver augmented with DASS at this 100-candidate budget approaches the original 1000-candidate accuracy while spending significantly less search time.

\paragraph{LaSR}

LaSR~\citep{lasr} augments the multi-population GP framework of PySR with an LLM-maintained natural-language concept library. In each generation, the LLM replaces mutation, crossover, or randomization operators with a small probability, using the current concepts as prompts. Elite expressions are abstracted into concepts and further evolved by the LLM to form the next set of concepts~\citep{holt2023dgsr, zhang2025rag}. We use 10 populations with 33 individuals per population, 20 outer iterations, and 550 GP cycles per outer iteration. The LLM replacement weights for crossover, mutation, and randomization are all set to $0.02$.

\paragraph{PySR}

PySR~\citep{pysr} is a multi-population island-GP framework~\citep{gp} that combines an annealing-style acceptance criterion with an expression-complexity penalty in the Pareto-front spirit of modern SR~\citep{cranmer2020discovering,virgolin2021improving}, with constants fitted by BFGS~\citep{bfgs}. PySR is a widely adopted baseline in recent SR benchmarks~\citep{lacava2021srbench,llmsrbench}, with parallel C++ alternatives such as Operon~\citep{burlacu2020operon}. In the main table, we use 15 populations with 33 individuals per population, 65 outer iterations, 550 GP cycles per outer iteration, and a maximum expression length of 30. The binary operator set is $\{+,-,\times,\div,\wedge\}$, and the unary operator set is $\{\exp,\log,\sqrt{\cdot},\sin,\cos,|\cdot|\}$. As shown in Section~\ref{4exp:budget}, PySR augmented with DASS at 65 outer iterations approaches the accuracy of the base solver run at 130 iterations at substantially lower search time.

\paragraph{GPlearn}
GPlearn~\citep{gplearn} is a scikit-learn-style single-population GP implementation that evolves expressions over a fixed population using tournament selection. We use a population of 500 individuals and a tournament size of 20, and run the evolution for 100 generations. The function set follows PySR's setting.

\paragraph{SPL}

SPL~\citep{spl} formulates symbolic regression as an MCTS~\citep{mcts} decision process over a context-free grammar. Each select--expand--simulate--backup cycle samples a grammar-derivation path, after which the resulting expression is fitted with constants using Powell optimization and assigned a reward; the reward is then backpropagated to update the UCB~\citep{ucb} estimates. SPL belongs to the broader family of neural-guided and search-based SR methods, including DSR~\citep{dsr}, neural-guided GP seeding~\citep{mundhenk2021seeding}, and the unified deep SR framework~\citep{landajuela2022udsr}. After several search episodes, a transplant operation is performed: substructures are extracted from the current elite expressions and introduced as new grammar rules. We run 10 transplant operations for each task, with each transplant consisting of $1000$ episodes. The maximum expression length is set to 20, the UCB decay coefficient to $\eta=0.999$, and the initial module capacity to $20$. As shown in Section~\ref{4exp:budget}, SPL augmented with DASS at 10 transplants achieves both higher accuracy and lower search time than the base solver run at 20 transplants.

\subsection{LLM Backbones}
\label{app:llm-backbones}

We use two open-weight LLMs as backbones. Qwen3.5-9B~\citep{qwen3.5} is a 9B-parameter post-trained model from the Qwen3.5 family, released as open weights in March 2026. Qwen3.5-9B uses a dense language backbone with a hybrid architecture
that interleaves Gated DeltaNet layers and gated attention layers,
together with a vision encoder. In our experiments, we use only
its text-generation capability.
On reasoning-oriented benchmarks such as MMLU-Pro and GPQA Diamond, Qwen3.5-9B matches or exceeds the previous-generation Qwen3-30B-A3B-Thinking-2507~\citep{qwen3technicalreport} model despite its smaller parameter count. Gemma-4-31B\footnote{
Model repository: \url{https://huggingface.co/google/gemma-4-31B}.} is Google DeepMind's 30.7B-parameter dense instruction-tuned Gemma 4 model, with a 256K-token context window and multilingual support.

Both models natively support configurable thinking modes. To align the inference paradigm with the non-reasoning LLMs used in the original LLM-SR and LaSR settings and to speed up inference, we explicitly disable thinking for both backbones. Both models are served through vLLM~\citep{kwon2023vllm}, with the sampling temperature set to $\tau=0.8$ and the maximum response length set to 8192 tokens.

\subsection{Compute Resources}
\label{app:compute}
All experiments are conducted on a single server equipped with
6$\times$ NVIDIA H800 PCIe GPUs (80\,GB each), dual AMD EPYC 9654
96-core CPUs (192 physical cores, 384 threads in total). LLM-based solvers (LLM-SR, LaSR) use the
GPUs for vLLM-based inference, while non-LLM solvers (PySR, GPlearn, SPL) run on the CPU. The average per-task search time for each
solver and domain is reported in the Time column of
Table~\ref{tab:main-results}.

\section{DASS Integration with Each Base Solver}
\label{app:dass-integration}

\begin{figure}[htbp]
  \centering
  \includegraphics[width=\linewidth]{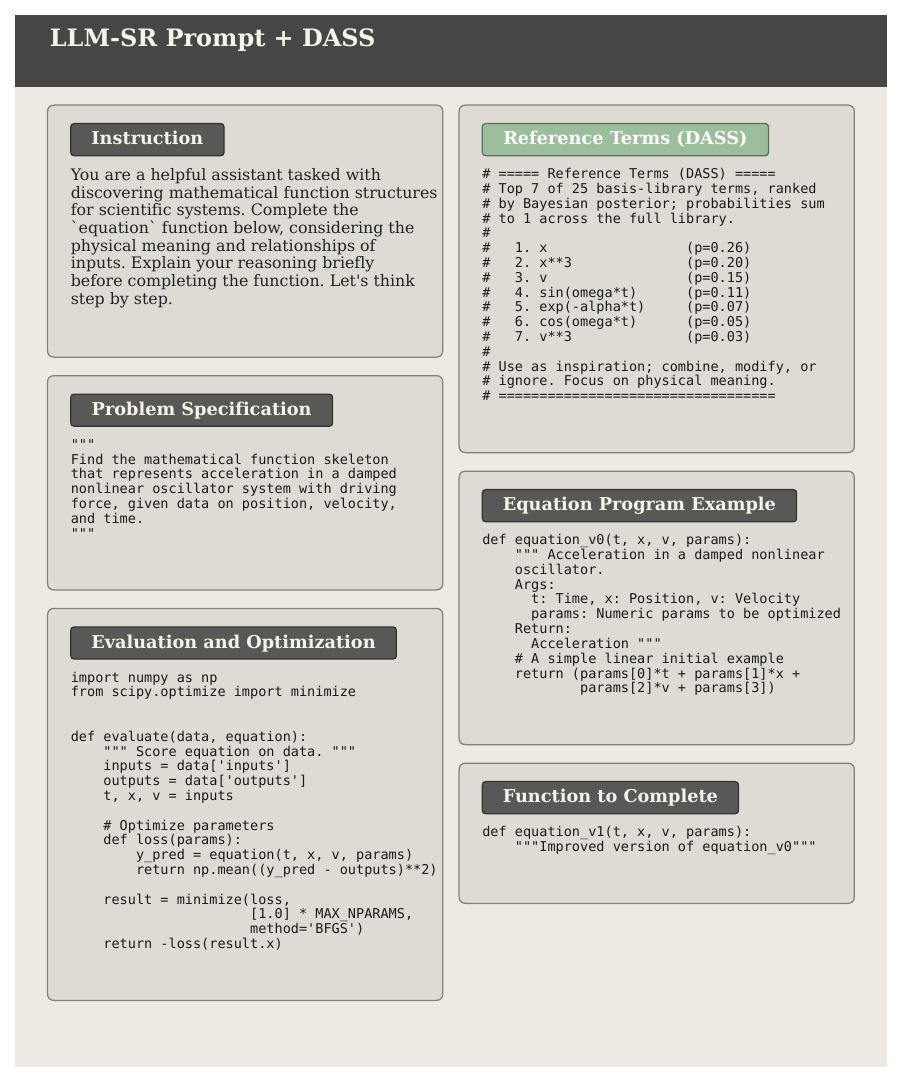}
  \caption{
      An example DASS-augmented LLM-SR prompt from the Physics subset of LLM-SRBench, shown in a compact two-column layout. The \emph{Reference Terms (DASS)} panel at the top of the second column is the only DASS-specific addition, listing the top-$k\%$ terms from $\mathcal{B}^{(t)}$ ranked by $\bm{\omega}^{(t)}$ as additional structural guidance for the LLM.
  }
  \label{fig:dass_prompt}
\end{figure}

\subsection{LLM-SR + DASS}
\label{app:integ-llmsr}

The search procedure of LLM-SR is fully preserved when combined with DASS: equation-skeleton proposal, constant fitting, candidate scoring, and cross-round search-state management all follow the original LLM-SR implementation. DASS is incorporated as an inserted module during prompt construction. Specifically, according to the current potential scores $\bm{\omega}^{(t)}$, the top-$k\%$ terms are selected from the basis set $\mathcal{B}^{(t)}$ and written into the ``Reference Terms'' section of the prompt. These terms are presented alongside the reference equations originally used by LLM-SR, serving as structural guidance for the language model. Fig.~\ref{fig:dass_prompt} illustrates the differences between the prompt sections in the original and DASS-augmented versions. The output protocol, sampling parameters, and evaluation procedure of the language model are left unchanged.

\subsection{LaSR + DASS}
\label{app:integ-lasr}

Unlike LLM-SR, which directly generates code, LaSR guides GP search using a natural-language concept library maintained by an LLM. Elite expressions extracted by GP are abstracted into concepts, which are then fed back in prompt form to the mutation and crossover operators in the next generation. The integration point of DASS with LaSR therefore lies at the input side of the concept library. At the DASS stage of round $t$, the top-$k\%$ terms are selected from $\mathcal{B}^{(t)}$ according to $\bm{\omega}^{(t)}$ and converted into natural-language concepts through a lightweight translation layer. These concepts are used as the initial concepts for the LaSR call in that round and are incorporated into the GP process together with the concepts evolved by LaSR itself. The internal GP algorithm of LaSR, including the population size, hall of fame, and mutation/crossover operators, as well as its concept-evolution mechanism, follows the original settings.

\subsection{PySR + DASS}
\label{app:integ-pysr}

PySR implements GP search in Julia and natively supports two mechanisms: guesses, which inject external expressions into the population, and warm start, which makes the next search continue from the population, complexity archive, and hall of fame produced by the previous search rather than restarting from scratch. Building on warm start, we divide the search for each task into several consecutive search calls, each referred to as a chunk. The entire process uses the same regressor instance. Before each chunk begins, DASS supplies PySR with two kinds of guesses: the top-$k\%$ terms from the current $(\mathcal{B}^{(t)}, \bm{\omega}^{(t)})$, and additive combinations of these terms, with combinations constrained to satisfy PySR's expression-complexity limits. Once inserted into the population, these guesses in PySR become ordinary GP individuals and can subsequently be selected, mutated, and crossed over under the same rules as other expressions; within that chunk, the search proceeds under PySR's default procedure.

\subsection{GPlearn + DASS}
\label{app:integ-gplearn}

GPlearn is a GP implementation with a scikit-learn-style interface. In each round of DASS, the top-$k\%$ terms from the current $(\mathcal{B}^{(t)}, \bm{\omega}^{(t)})$, together with additive combinations of these terms constrained to satisfy GPlearn's program-depth limits, are compiled into GPlearn's internal program objects and used to replace the same number of individuals in the current population, thereby keeping the population size unchanged. When a new generation starts, GPlearn's tournament selection draws a fixed-size random subset from the entire population and picks the highest-fitness individual as the parent. The injected individuals and the original individuals participate in this selection process under the same rules. Across multiple injections, GPlearn continues training on the same regressor instance, with the generation counter accumulated and the population, mutation and crossover operators, parsimony coefficient, and other settings all retained.

\subsection{SPL + DASS}
\label{app:integ-spl}

SPL constructs expression trees over a context-free grammar (CFG)~\citep{cfg} using MCTS. Each action in the search corresponds to selecting a grammar production rule, which expands the current stack-top nonterminal into a more specific substructure. Once a complete expression is obtained, its constant placeholders are numerically fitted, and the node estimates are updated through UCB backpropagation. The injection point of DASS in SPL is the production-rule set. At the DASS stage of round $t$, the top-$k\%$ terms in $\mathcal{B}^{(t)}$, ranked by $\bm{\omega}^{(t)}$, are each encapsulated as a new production rule. These rules are added to the grammar alongside SPL's base production rules and the rules generated by transplantation, thereby introducing effective signals from DASS into the search process. The MCTS selection, expansion, simulation, and backpropagation loop, the UCB scoring form, the transplant triggering mechanism, and the maintenance of high-reward module rules all follow the original SPL settings.

\section{Extensions to Solvers Without Explicit Prior-Injection Interfaces}
\label{Ext}

\begin{table}[htbp]
  \centering
  \caption{Performance of DSR and GenSR before and after coupling with DASS through the action-prior (DSR) and latent-prior (GenSR) interfaces. DASS-augmented rows shaded gray.}
  \label{tab:extensions}
  \resizebox{\linewidth}{!}{\begingroup
\def\g{\cellcolor{gray!10}}%
\begin{tabular}{llcccccccc}
\toprule
\multirow{2}[4]{*}{\textbf{Category}}
  & \multirow{2}[4]{*}{\textbf{Method}}
  & \multicolumn{2}{c}{\textbf{Phys}}
  & \multicolumn{2}{c}{\textbf{MatSci}}
  & \multicolumn{2}{c}{\textbf{Chem}}
  & \multicolumn{2}{c}{\textbf{Bio}} \\
\cmidrule(lr){3-4}\cmidrule(lr){5-6}\cmidrule(lr){7-8}\cmidrule(lr){9-10}
 & & NMSE$\downarrow$ & $\mathrm{Acc}_{0.01}$$\uparrow$
   & NMSE$\downarrow$ & $\mathrm{Acc}_{0.01}$$\uparrow$
   & NMSE$\downarrow$ & $\mathrm{Acc}_{0.01}$$\uparrow$
   & NMSE$\downarrow$ & $\mathrm{Acc}_{0.01}$$\uparrow$ \\
\midrule
\multirow{2}{*}{\textbf{RL-based}}
 & DSR                          & 4.82E-01 & 0   & 8.65E-03 & 0    & 9.31E-01 & 0    & 1.54E+00 & 0   \\
 & \g\textbf{DSR+DASS}          & \g\textbf{8.41E-02} & \g 0 & \g\textbf{2.97E-03}  & \g 0 & \g\textbf{2.04E-01} & \g 0 & \g\textbf{7.18E-01} & \g 0 \\
\midrule
\multirow{2}{*}{\textbf{Latent-space-based}}
 & GenSR                        & 5.21E-03 & 4.5 & 8.32E-05 & 18.7 & 2.86E-01 & 8.3  & 7.85E-01 & 4.2 \\
 & \g\textbf{GenSR+DASS}        & \g\textbf{1.34E-03} & \g\textbf{12.1} & \g\textbf{2.18E-05} & \g\textbf{33.3} & \g\textbf{4.21E-02} & \g\textbf{16.7} & \g\textbf{8.62E-02} & \g\textbf{9.7} \\
\bottomrule
\end{tabular}%
\endgroup}
\end{table}

For some SR methods, DASS cannot be integrated in a purely plug-and-play manner because the solver does not expose an explicit interface for injecting symbolic priors or search biases. Nevertheless, DASS can still be coupled to such methods by intervening at the latent-search level or at the policy level. Below we discuss two representative cases.

\subsection{Coupling DASS with latent-space-based SR solver (e.g., GenSR)}

GenSR~\citep{li2026gensr} belongs to the broader family of neural SR approaches that operate in representations, including equation-learner networks~\citep{sahoo2018eql} and end-to-end transformer-based SR~\citep{e2e, meidani2023snip, dascoli2024odeformer}. Specifically, GenSR performs SR by first constructing a continuous generative latent space of equations using a dual-branch CVAE, then using the prior branch to localize a promising latent region, and finally refining the search by a modified CMA-ES procedure in latent space. Since GenSR does not directly manipulate basis terms during search, DASS cannot be injected into it through a standard symbolic-prior interface. Instead, DASS can be coupled to GenSR by providing a latent initialization prior derived from the current auxiliary subspace.

One natural integration strategy is to use DASS as a latent initialization prior. Specifically, we first sample a set of high-potential expressions from the calibrated DASS subspace and encode them into the GenSR latent space, obtaining latent anchors $\{z_{\rm DASS}\}$. These anchors define a DASS-induced latent search signal $\mathcal{N}(\mu_{\rm DASS},\Sigma_{\rm DASS})$. This signal can be fused with the original latent initialization by convex interpolation as Eq.~\ref{gensr}. Here, $(\mu_{\rm base},\Sigma_{\rm base})$ are the original encoder outputs of GenSR, $\alpha\in[0,1]$ controls the relative strength of the original latent prior and the DASS-induced structural prior. After GenSR decodes and evaluates candidate equations, its high-quality solutions are fed back to DASS exactly as in our standard framework: newly discovered useful terms are extracted from these decoded equations, low-utility basis terms are replaced, and the updated basis term set is used to construct the DASS-induced prior for the next round. In this way, DASS does not reshape the latent space of GenSR itself; rather, it reshapes its own term-defined auxiliary subspace, which in turn continuously biases the latent initialization of GenSR.
\begin{equation}\label{gensr}
    \mu'=\alpha\mu_{\rm base}+(1-\alpha)\mu_{\rm DASS},\quad\Sigma'=\alpha\Sigma_{\rm base}+(1-\alpha)\Sigma_{\rm DASS}.
\end{equation}

\subsection{Coupling DASS with policy-based SR solver (e.g., DeepSR)}
DeepSR~\citep{dsr} formulates SR as learning an autoregressive policy over expression tokens: an RNN emits a distribution over tokens in prefix order, sampled expressions are instantiated and evaluated, and the policy is updated by a risk-seeking policy gradient objective. We integrate DASS through a structural prior over policy actions.

Let $p_\theta(\tau_i|\tau_{<i})$ denote the original token policy of DeepSR at step $i$, where $\tau_i$ is the next token and $\tau_{<i}$ is the previously generated prefix. To define a DASS-induced action prior, we first represent each basis term $s\in\mathcal{B}^{(t)}$ by its prefix-order token sequence $\pi(s)$. For a candidate next token $v$, we define its compatibility with the current DASS subspace as
$$
b^{(t)}(v\mid \tau_{<i})
=
\frac{\sum_{s\in \mathcal{B}^{(t)}} \omega^{(t)}(s)\,\phi_s(\tau_{<i},v)}
{\sum_{s\in \mathcal{B}^{(t)}} \omega^{(t)}(s)+\epsilon},
$$
where $\phi_s(\tau_{<i},v)=1$ if appending $v$ to the current prefix keeps the partial sequence compatible with the prefix of $\pi(s)$, and $\phi_s(\tau_{<i},v)=0$ otherwise. Intuitively, $b^{(t)}(v\mid \tau_{<i})$ is large when selecting $v$ makes the current partial expression more consistent with high-potential basis terms in the current DASS subspace.

The DASS-guided token policy is then defined as
$$
p_{\theta}^{\mathrm{DASS}}(\tau_i=v\mid \tau_{<i})
\propto
p_{\theta}(\tau_i=v\mid \tau_{<i})
\exp\!\big(\lambda\, b^{(t)}(v\mid \tau_{<i})\big)
$$
where $\lambda\geq0$ controls the strength of structural guidance. This formulation preserves the original DeepSR policy while biasing sampling toward tokens that are more compatible with the current auxiliary subspace.

The resulting sampled expressions are still evaluated by the original DeepSR objective, and the policy is updated in its native reinforcement-learning manner. To retain the DASS closed loop, high-reward expressions generated by the policy are further treated as solver feedback: useful new terms are extracted, the basis term set is reshaped, and the updated $(\mathcal{B}^{(t+1)},\omega^{(t+1)})$ induces the action prior in the next DASS round.

\section{Symbolic Accuracy and Expression Complexity}
\label{app:dass-symbolic-accuracy}
Numerical accuracy and symbolic recovery assess different aspects of symbolic regression. We therefore additionally report symbolic accuracy (SA), assessed using the two-stage adjudication protocol of SR-Scientist~\citep{xia2025sr}, and expression complexity following the SRBench convention. All accuracy values in the following aggregate tables are percentages.

\begin{table}[tbp]
\centering
\small
\caption{Numerical accuracy, expression complexity, and symbolic accuracy on LSR-Synth for LLM-based solvers. DASS variants are compared with their base solvers at the same downstream search budget. The higher-budget LLM-SR result is provided separately as a reference. Accuracy values are percentages; higher accuracy and lower complexity are preferred.}
\label{tab:dass-sa-llm}
\setlength{\tabcolsep}{5pt}
\begin{tabular}{llrrr}
\toprule
Method & Budget & $\mathrm{Acc}_{0.01}$ & Complexity & SA \\
\midrule
\multicolumn{5}{l}{\textit{Qwen3.5-9B}} \\
LLM-SR & 100 calls & 13.95 & 48.5 & 2.58 \\
LLM-SR + DASS & 100 calls & 34.85 & 50.1 & 4.65 \\
LLM-SR (higher budget) & 1,000 calls & 35.60 & 55.0 & 4.91 \\
LaSR & 20 iterations & 10.58 & 24.2 & 3.10 \\
LaSR + DASS & 20 iterations & 21.94 & 22.7 & 4.91 \\
\midrule
\multicolumn{5}{l}{\textit{Gemma-4-31B}} \\
LLM-SR & 100 calls & 20.39 & 47.6 & 3.36 \\
LLM-SR + DASS & 100 calls & 36.20 & 45.2 & 5.17 \\
LaSR & 20 iterations & 15.00 & 23.4 & 2.58 \\
LaSR + DASS & 20 iterations & 25.83 & 21.9 & 5.17 \\
\bottomrule
\end{tabular}
\end{table}

\begin{table}[tbp]
\centering
\small
\caption{Additional evaluation of GP-based and MCTS-based solvers on LSR-Synth. Higher-budget base-solver configurations are reference points rather than matched-budget comparisons. Accuracy values are percentages.}
\label{tab:dass-sa-nonllm}
\setlength{\tabcolsep}{5pt}
\begin{tabular}{llrrr}
\toprule
Method & Budget & $\mathrm{Acc}_{0.01}$ & Complexity & SA \\
\midrule
PySR & 65 iterations & 17.83 & 19.6 & 3.10 \\
PySR + DASS & 65 iterations & 27.89 & 21.2 & 3.88 \\
PySR (higher budget) & 130 iterations & 27.40 & 23.8 & 4.39 \\
\midrule
SPL & 10 transplants & 12.12 & 16.8 & 0.52 \\
SPL + DASS & 10 transplants & 21.73 & 20.4 & 0.52 \\
SPL (higher budget) & 20 transplants & 18.10 & 18.6 & 0.78 \\
\midrule
gplearn & 100 generations & 0.00 & 65.1 & 0.00 \\
gplearn + DASS & 100 generations & 0.78 & 58.7 & 0.00 \\
\bottomrule
\end{tabular}
\end{table}

Tables~\ref{tab:dass-sa-llm} and~\ref{tab:dass-sa-nonllm} show that DASS improves SA for LLM-SR, LaSR, and PySR at matched downstream budgets. For example, with Qwen3.5-9B, LLM-SR improves from $2.58\%$ to $4.65\%$ SA, while LaSR improves from $3.10\%$ to $4.91\%$. These gains are smaller than the corresponding gains in accuracy-to-tolerance. For the expression complexity: it increases for Qwen3.5-9B LLM-SR, PySR, and SPL, but decreases for both LaSR configurations, Gemma-4-31B LLM-SR, and gplearn.

\section{Integration with SR-Scientist and SGA}
\label{app:dass-additional-solvers}

We further evaluate DASS with SR-Scientist~\citep{xia2025sr} and SGA~\citep{ma2024llm}. Since DASS is an auxiliary search strategy, the matched comparisons below measure its incremental benefit within each solver.

\paragraph{SR-Scientist.}
A simple integration is to append a small set of basis terms, sampled according to their DASS potential scores, to the agent’s context at the beginning of each outer iteration, alongside the high-performing equations retrieved from its experience buffer. These terms serve as optional structural hints: the agent may combine, modify, or ignore them while using its original data-analysis and equation-evaluation tools. The guidance remains fixed throughout the inner interaction trajectory. At the end of the outer iteration, high-quality evaluated equations are returned to DASS for term extraction and subspace reshaping. Subsequent iterations receive the updated guidance following the scheduled DASS calibration.

We use GPT-OSS-120B and a maximum downstream budget of $10\times25$ calls for both the base solver and its DASS-enhanced variant. Table~\ref{tab:dass-srscientist} reports two numerical tolerances together with complexity and SA. DASS increases $\mathrm{Acc}_{0.01}$ by $11.63$ percentage points and $\mathrm{Acc}_{0.001}$ by $3.88$ points. SA increases from $4.65\%$ to $6.20\%$, while complexity decreases from $28.5$ to $26.2$. Thus, the paired results show improvements in both numerical accuracy and symbolic recovery without an increase in average expression complexity.

\begin{table}[tbp]
\centering
\small
\caption{Integration with SR-Scientist using GPT-OSS-120B on LSR-Synth. Both configurations use a maximum budget of $10\times25$ downstream calls. Accuracy values are percentages.}
\label{tab:dass-srscientist}
\begin{tabular}{lrrrr}
\toprule
Method & $\mathrm{Acc}_{0.01}$ & $\mathrm{Acc}_{0.001}$ & Complexity & SA \\
\midrule
SR-Scientist & 49.61 & 39.53 & 28.5 & 4.65 \\
SR-Scientist + DASS & 61.24 & 43.41 & 26.2 & 6.20 \\
\bottomrule
\end{tabular}
\end{table}

\paragraph{SGA.}
DASS can be integrated into SGA’s outer-level expression-generation procedure by adding a small set of potential-score-sampled basis terms to the prompt, alongside existing high-performing candidates and evaluation feedback. The terms provide optional building blocks for proposing or revising expressions. Each generated candidate then undergoes SGA’s original inner-level parameter optimization and evaluation. After each search round, high-quality evaluated expressions are returned to DASS to extract new terms and reshape the auxiliary subspace. The refreshed guidance is used in subsequent prompts, while the parameter optimizer, candidate-ranking rule, and exploration settings remain unchanged.

We evaluate both configurations with 100 downstream calls under each backbone. As shown in Table~\ref{tab:dass-sga}, DASS improves $\mathrm{Acc}_{0.01}$ by $10.34$ percentage points with Qwen3.5-9B and by $9.56$ points with Gemma-4-31B. SA increases by $1.55$ points under both backbones. Complexity increases slightly with Qwen3.5-9B ($40.4$ to $41.5$) and decreases with Gemma-4-31B ($39.6$ to $36.8$).

\begin{table}[tbp]
\centering
\small
\caption{Integration with SGA on LSR-Synth using 100 downstream calls. Each comparison holds the backbone and downstream call budget fixed. Accuracy values are percentages.}
\label{tab:dass-sga}
\setlength{\tabcolsep}{5pt}
\begin{tabular}{llrrr}
\toprule
Backbone & Method & $\mathrm{Acc}_{0.01}$ & Complexity & SA \\
\midrule
Qwen3.5-9B & SGA & 9.30 & 40.4 & 1.81 \\
Qwen3.5-9B & SGA + DASS & 19.64 & 41.5 & 3.36 \\
\midrule
Gemma-4-31B & SGA & 13.70 & 39.6 & 2.58 \\
Gemma-4-31B & SGA + DASS & 23.26 & 36.8 & 4.13 \\
\bottomrule
\end{tabular}
\end{table}

\section{Sensitivity to the Initial Basis Library}
\label{app:dass-initialization}

We study initialization using LLM-SR and PySR, with initial basis libraries generated by Llama-3.1-8B, GPT-4o, or Gemma-4-31B. The initializer receives only domain context, semantic descriptions of the input and output variables, and formatting constraints. It does not receive task-specific ground-truth equations, training or test values, or the candidate terms used to construct LSR-Synth.

For each generator, we evaluate a randomized initialization: terms are randomly selected from that generator's candidate pool instead of being ranked by generation frequency, and their initial potential scores are randomly assigned and normalized. The downstream solver configuration, search budget, basis-set size, number of DASS update rounds, and remaining hyperparameters are held fixed. The column labels in Table~\ref{tab:dass-initialization} refer to the initialization model.

\begin{table}[tbp]
\centering
\small
\caption{Sensitivity to the initial basis library on LSR-Synth, measured by $\mathrm{Acc}_{0.01}$ (\%). Baselines do not use a basis library; their results are repeated for comparison. ``Randomized'' changes both term selection and initial potential scores within an LLM-generated candidate pool.}
\label{tab:dass-initialization}
\setlength{\tabcolsep}{4pt}
\begin{tabular}{lrrr}
\toprule
Method & Llama-3.1-8B & GPT-4o & Gemma-4-31B \\
\midrule
LLM-SR & 13.95 & 13.95 & 13.95 \\
LLM-SR + DASS & 25.58 & 34.85 & 37.21 \\
LLM-SR + DASS (randomized) & 22.48 & 31.01 & 33.33 \\
\midrule
PySR & 17.83 & 17.83 & 17.83 \\
PySR + DASS & 23.26 & 27.89 & 30.23 \\
PySR + DASS (randomized) & 20.16 & 25.58 & 28.68 \\
\bottomrule
\end{tabular}
\end{table}

DASS improves over the corresponding base solver under every tested initialization, including randomized initialization. However, initialization quality has a measurable effect: standard DASS ranges from $25.58\%$ to $37.21\%$ for LLM-SR and from $23.26\%$ to $30.23\%$ for PySR across generators. Randomization reduces accuracy by $3.10$--$3.88$ percentage points for LLM-SR and by $1.55$--$3.10$ points for PySR. These results support robustness to the choice of initializer, but not insensitivity to initialization quality. The results show that calibration and reshaping mitigating a weaker starting point under a finite update budget.

\section{Transferability of the Learned Search Space}
\label{app:dass-transfer}

We evaluate whether a learned auxiliary search space can initialize a different task using low-budget PySR on the 44 Physics tasks. For within-domain transfer, each target task is initialized using the final search space learned on a randomly selected different Physics task. For cross-domain transfer, the source space is learned on a Chemistry task, and variables are randomly mapped between source and target tasks. DASS continues to calibrate and reshape the transferred space on the target task; this experiment therefore evaluates transfer as initialization rather than a frozen or zero-shot search space. The remaining settings follow the main experiments.

\begin{table}[tbp]
\centering
\small
\caption{Transfer to Physics tasks with low-budget PySR. All DASS settings continue adapting the space on the target task. The reported $\mathrm{Acc}_{0.01}$ values are percentages.}
\label{tab:dass-transfer}
\begin{tabular}{lr}
\toprule
Initial search space for the target task & $\mathrm{Acc}_{0.01}$ \\
\midrule
No auxiliary space (PySR without DASS) & 18.2 \\
Physics domain-level prior (standard DASS) & 22.7 \\
Final space from another Physics task & 24.2 \\
Final space from a Chemistry task & 20.5 \\
\bottomrule
\end{tabular}
\end{table}

Within-domain transfer improves $\mathrm{Acc}_{0.01}$ by $1.5$ percentage points over standard DASS (Table~\ref{tab:dass-transfer}). Nevertheless, task-level NMSE degrades on 15 of the 44 tasks, showing that a better aggregate success rate does not imply uniformly improved fits. Chemistry-to-Physics transfer exceeds PySR without DASS by $2.3$ points but falls below the Physics domain-level initialization by $2.2$ points. The results suggest that source--target compatibility matters and that transferring a learned space can introduce negative transfer relative to a suitable domain prior. Because target-task adaptation remains active, the cross-domain gain cannot be attributed to the transferred space alone.

\section{Term-Level Utility and Interaction-Aware Attribution}
\label{app:dass-attribution}

We compare our term-utility estimator in the main text with two alternatives. The \emph{counting} variant adds one unit of utility whenever a term occurs in an elite expression. The \emph{marginal} variant replaces the fitness numerator in the standard utility estimator with
\begin{equation}
\Delta_s F(f)=\max\!\left\{0,\,F(f)-F(f\setminus s)\right\},
\label{eq:dass-marginal-credit}
\end{equation}
where $F(f)=\mathrm{Fitness}(f)$ and $f\setminus s$ denotes the expression after removing term $s$. The coefficients of the reduced expression are refitted before its fitness is evaluated. This variant measures the effect of removing a term in the context of the remaining terms, while retaining the expression-length normalization and the subsequent DASS update procedure. It captures some context dependence, but is not a complete decomposition of higher-order interactions.

\begin{table}[tbp]
\centering
\small
\caption{Term-level utility attribution with LLM-SR and a downstream budget of 100 calls. Accuracy is $\mathrm{Acc}_{0.01}$ (\%); time is the reported search time in seconds. The marginal variant includes additional term-removal, coefficient-refitting, and fitness-evaluation work.}
\label{tab:dass-attribution}
\begin{tabular}{lrr}
\toprule
Method & $\mathrm{Acc}_{0.01}$ & Time (s) \\
\midrule
LLM-SR & 13.95 & 187.6 \\
LLM-SR + Counting DASS & 25.84 & 241.5 \\
LLM-SR + Original DASS & 34.85 & 245.3 \\
LLM-SR + Marginal DASS & 36.18 & 364.1 \\
\bottomrule
\end{tabular}
\end{table}

Table~\ref{tab:dass-attribution} shows that the original estimator improves accuracy by $9.01$ percentage points over counting, with only $3.8$ additional seconds of search time. Incorporating fitness information is therefore more effective in this setting than counting elite occurrences alone. Marginal attribution yields a further $1.33$-point gain over the original estimator, but increases time from $245.3$ to $364.1$ seconds, approximately $48.4\%$. This additional cost is consistent with the need to refit and reevaluate reduced expressions. The original estimator offers a favorable observed accuracy--time trade-off, while marginal attribution is a useful alternative when additional computation is available. 

\section{Representative Recovered Expressions}
\label{app:dass-cases}

We report four representative tasks spanning Physics, Biology, Chemistry, and Materials Science, using PySR and LLM-SR. In the following examples, GT denotes the ground truth, Base denotes the base-solver output, and DASS denotes its DASS-enhanced output. The accuracy and SA labels refer only to the DASS output. Coefficients are rounded to four decimal places for display; the reported success labels apply to the evaluated expressions rather than a fresh evaluation of these rounded transcriptions.

\paragraph{PO4: recovery of a missing nonlinear term.}
For this Physics task, PySR + DASS is successful under both $\mathrm{Acc}_{0.01}$ and SA:
\begin{align*}
\text{GT:}\quad \dot v(t)&=0.1063\sin(t)-0.4844\sin(v)-0.9688v,\\
\text{Base:}\quad \dot v(t)&=0.1063\sin(t)-1.4530v,\\
\text{DASS:}\quad \dot v(t)&=0.1063\sin(t)-0.4843\sin(v)-0.9689v.
\end{align*}
The base solver absorbs the state-dependent sinusoidal term into a linear term. DASS recovers both the sinusoidal and linear components, with small differences in the displayed coefficients.

\paragraph{BPG13: recovery of the population-growth structure.}
For this Biology task, LLM-SR + DASS is successful under both $\mathrm{Acc}_{0.01}$ and SA:
\begin{align*}
\text{GT:}\quad \dot P(t)
&=0.8770P\sin(0.5672t)+0.7014(1-P/65.7541)P,\\
\text{Base:}\quad \dot P(t)
&=-4.6339\!\times\!10^{-5}P^2t+0.0020P^2+0.0055Pt\\
&\quad-0.0391P\sin(t)+0.0107P\cos(t)-0.2498P\\
&\quad+51.4725P\exp(-t)-0.0215t^2+1.1205t\\
&\quad-100.6531t\exp(-t)+0.4811\sin(t)\\
&\quad-2.3039\cos(t)-9.5515+93.2485\exp(-t),\\
\text{DASS:}\quad \dot P(t)
&=0.8770P\sin(0.5672t)+0.7014(1-P/65.7541)P.
\end{align*}
The base expression combines numerous polynomial, trigonometric, and exponential terms. The DASS-enhanced output matches the displayed ground-truth form, recovering the periodic modulation and logistic-growth component.

\paragraph{CKR19: an accurate polynomial surrogate without symbolic recovery.}
For this Chemistry task, LLM-SR + DASS satisfies $\mathrm{Acc}_{0.01}$ but not SA:
\begin{align*}
\text{GT:}\quad \dot A(t)&=-0.5637A^2+0.1314\sin(0.7617A),\\
\text{Base:}\quad \dot A(t)&=-0.5801A^2-0.0004At+0.1082A\\
&\quad+0.0001t-0.0008,\\
\text{DASS:}\quad \dot A(t)&=-0.5637A^2+0.1001A-0.0097A^3+0.0003A^5.
\end{align*}
The DASS-enhanced output retains the quadratic term but replaces the sine term with an odd polynomial. Such a surrogate can satisfy a numerical tolerance over the evaluated range without being symbolically equivalent to the target function.

\paragraph{MatSci13: numerical success with a different functional form.}
For this Materials Science task, PySR + DASS satisfies $\mathrm{Acc}_{0.01}$ but not SA:
\begin{align*}
\text{GT:}\quad \sigma(\epsilon,T)
&=3.6246\epsilon[-0.0590(T-367.0230)+1]\\
&\quad+94.2050\epsilon^{0.4475}\exp\!\left(-\frac{55.0161}{8.3140T}\right)\\
&\quad+5.1243\epsilon\exp[-(T-367.0230)^2],\\
\text{Base:}\quad \sigma(\epsilon,T)
&=104.0935T\epsilon+0.0306T+10418.8006\epsilon^2\\
&\quad-4303.7112\epsilon\sqrt{T+1}-13424.8524\epsilon\\
&\quad+\epsilon\exp(-T)+0.0176\sin(T)+56657.7951\tanh(\epsilon),\\
\text{DASS:}\quad \sigma(\epsilon,T)
&=-0.1600T\epsilon-50.6829\epsilon+218.5383\sqrt{\epsilon}\\
&\quad-25145.6269\frac{\sqrt{\epsilon}}{T}.
\end{align*}
The DASS-enhanced expression is shorter than the base output, yet it omits the target's exponential temperature dependence and uses a different power of strain. 

The four examples distinguish improved numerical fitting from recovery of the target expression. The aggregate results show that DASS improves SA in some solver configurations, but the gains are limited and absolute SA remains low. DASS is an auxiliary strategy intended to accelerate and guide an existing solver’s search; it does not independently generate and optimize the final solution or guarantee symbolic recovery. Final performance depends jointly on the guidance and the base solver’s expression representation, proposal mechanism, parameter fitting, and candidate selection under the available budget. DASS can help a solver explore useful structures, but its guidance does not eliminate the solver’s limitations or the difficulty of distinguishing a numerically accurate surrogate from the correct symbolic expression.

\section{Additional Experimental Results}
\label{app:additional-results}

\subsection{Out-of-Distribution Performance}
\label{app:ood}

All tasks in LSR-Synth provide both ID and OOD test sets. The OOD test set extends the range of at least one variable (typically time) beyond the training interval (see Appendix~\ref{app:dataset}), and is used to evaluate the stability of the fitted expression under variable extrapolation. Table~\ref{tab:ood-results} reports the corresponding OOD performance under the same configuration as the main table in Section~\ref{4exp:mainres} (Table~\ref{tab:main-results}), with only the Time column omitted because the search time is identical.

For every (base solver, domain) configuration, adding DASS reduces NMSE. $\mathrm{Acc}_{0.01}$ also improves in most configurations, with the only exceptions being GPlearn on Physics, Chemistry, Biology, which remain at 0. In these cases, the OOD errors of the base solver are already far above the accuracy threshold; although DASS substantially reduces NMSE, the improvement is still insufficient to cross the threshold, and therefore does not indicate a performance regression. For most methods, when moving from ID to OOD after incorporating DASS, $\mathrm{Acc}_{0.01}$ decreases slightly on Physics, Chemistry, and Biology. On Material Science, however, the opposite pattern is generally observed, with OOD $\mathrm{Acc}_{0.01}$ exceeding ID $\mathrm{Acc}_{0.01}$; this overall trend is consistent with the observations of \citet{xia2025sr} on LSR-Synth. The stable gains of DASS on OOD evaluation indicate that the injected structural priors help the base solver recover the true functional form of the expression, rather than overfitting to specific numerical patterns in the ID test points.

\begin{table}[htbp]
  \centering
  \caption{
    OOD performance across 4 LSR-Synth domains. Time is omitted (identical to Table~\ref{tab:main-results}); DASS-augmented rows are shaded gray.
  }
  \label{tab:ood-results}
  \resizebox{\linewidth}{!}{\begingroup
\def\g{\cellcolor{gray!10}}%
\begin{tabular}{llcccccccc}
\toprule
\multirow{2}[4]{*}{\textbf{Category}}
  & \multirow{2}[4]{*}{\textbf{Method}}
  & \multicolumn{2}{c}{\textbf{Phys}}
  & \multicolumn{2}{c}{\textbf{MatSci}}
  & \multicolumn{2}{c}{\textbf{Chem}}
  & \multicolumn{2}{c}{\textbf{Bio}} \\
\cmidrule(lr){3-4}\cmidrule(lr){5-6}\cmidrule(lr){7-8}\cmidrule(lr){9-10}
 & & NMSE$\downarrow$ & $\mathrm{Acc}_{0.01}$$\uparrow$
   & NMSE$\downarrow$ & $\mathrm{Acc}_{0.01}$$\uparrow$
   & NMSE$\downarrow$ & $\mathrm{Acc}_{0.01}$$\uparrow$
   & NMSE$\downarrow$ & $\mathrm{Acc}_{0.01}$$\uparrow$ \\
\midrule
\multirow{10}{*}[-10pt]{\textbf{LLM-based}}
 & \multicolumn{9}{c}{Qwen3.5-9B} \\
\cmidrule(lr){2-10}
 & LLMSR                      & 2.18E-02 & 8.3 & 1.10E-05 & 45.3 & 7.26E-03 & 5.6 & 4.46E-02 & 5.6 \\
 & \g\textbf{LLMSR+DASS}    & \g\textbf{2.07E-03} & \g\textbf{28.0} & \g\textbf{2.32E-06} & \g\textbf{85.3} & \g\textbf{1.84E-03} & \g\textbf{22.2} & \g\textbf{7.33E-03} & \g\textbf{12.5} \\
 & LASR                       & 9.90E-02 & 5.3 & 5.36E-04 & 30.7 & 9.54E-02 & 10.2 & 1.10E-03 & 4.2 \\
 & \g\textbf{LASR+DASS}    & \g\textbf{1.19E-02} & \g\textbf{19.7} & \g\textbf{6.78E-05} & \g\textbf{50.7} & \g\textbf{1.09E-02} & \g\textbf{16.7} & \g\textbf{8.21E-04} & \g\textbf{11.1} \\
\cmidrule(lr){2-10}
 & \multicolumn{9}{c}{Gemma-4-31B} \\
\cmidrule(lr){2-10}
 & LLMSR                      & 8.07E-03 & 12.1 & 1.16E-05 & 50.7 & 2.46E-03 & 14.8 & 2.08E-02 & 8.3 \\
 & \g\textbf{LLMSR+DASS}    & \g\textbf{1.01E-03} & \g\textbf{34.8} & \g\textbf{9.49E-06} & \g\textbf{70.7} & \g\textbf{9.88E-04} & \g\textbf{26.9} & \g\textbf{3.08E-03} & \g\textbf{16.7} \\
 & LASR                       & 4.98E-02 & 9.8 & 3.02E-04 & 38.7 & 5.12E-02 & 13.9 & 2.55E-01 & 5.6 \\
 & \g\textbf{LASR+DASS}    & \g\textbf{7.67E-03} & \g\textbf{22.0} & \g\textbf{3.12E-05} & \g\textbf{53.3} & \g\textbf{6.51E-03} & \g\textbf{22.2} & \g\textbf{5.11E-02} & \g\textbf{12.5} \\
\midrule
\multirow{4}{*}{\textbf{GP-based}}
 & PySR                       & 1.89E-02 & 19.7 & 6.76E-05 & 52.0 & 2.15E-02 & 9.3 & 1.19E-01 & 16.7 \\
 & \g\textbf{PySR+DASS}    & \g\textbf{5.12E-03} & \g\textbf{24.2} & \g\textbf{1.64E-05} & \g\textbf{56.0} & \g\textbf{6.85E-03} & \g\textbf{24.1} & \g\textbf{1.98E-02} & \g\textbf{30.6} \\
 & GPlearn                    & 5.36E-01 & 0.0 & 1.01E-02 & 2.7 & 1.02E+00 & 0.0 & 9.26E-01 & 0.0 \\
 & \g\textbf{GPlearn+DASS}    & \g\textbf{9.30E-02} & \g 0.0 & \g\textbf{3.52E-03} & \g\textbf{5.3} & \g\textbf{2.55E-01} & \g 0.0 & \g\textbf{5.34E-01} & \g 0.0 \\
\midrule
\multirow{2}{*}{\textbf{MCTS-based}}
 & SPL                        & 4.09E-02 & 5.3 & 1.84E-04 & 44.0 & 2.33E-01 & 11.1 & 6.14E-01 & 5.6 \\
 & \g\textbf{SPL+DASS}    & \g\textbf{1.01E-02} & \g\textbf{21.2} & \g\textbf{3.24E-05} & \g\textbf{48.0} & \g\textbf{3.25E-02} & \g\textbf{22.2} & \g\textbf{4.84E-02} & \g\textbf{11.1} \\
\bottomrule
\end{tabular}%
\endgroup
}
\end{table}

\subsection{Per-Domain View of Low-Budget DASS vs. High-Budget Base}
\label{app:per-domain-budget}

\begin{figure}[htbp]
  \centering
  \includegraphics[width=\linewidth]{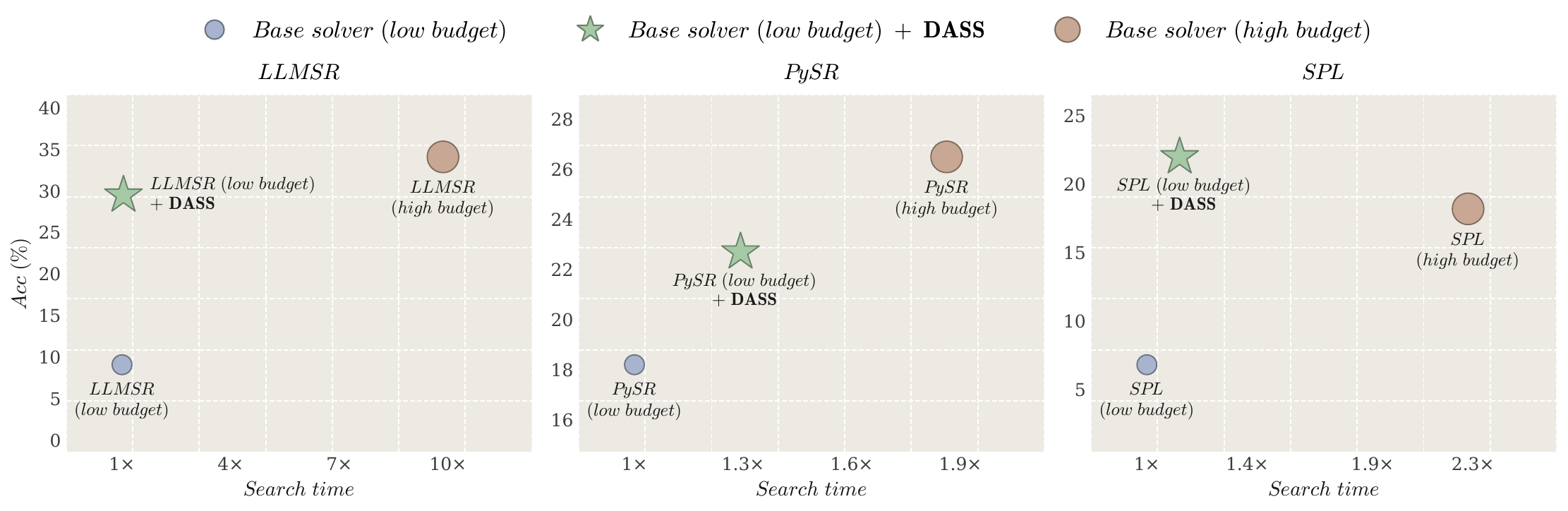}
  \caption{
    Budget efficiency on the Physics domain. Each panel corresponds to one base solver (LLMSR, PySR, SPL). The horizontal axis shows relative search time (base-solver low budget set to $1\times$); the vertical axis shows $\mathrm{Acc}_{0.01}$. The three markers within each panel correspond to the base solver at low budget, the base solver augmented with DASS at low budget, and the base solver at high budget.
  }
  \label{fig:phys_budget}
\end{figure}

\begin{figure}[htbp]
  \centering
  \includegraphics[width=\linewidth]{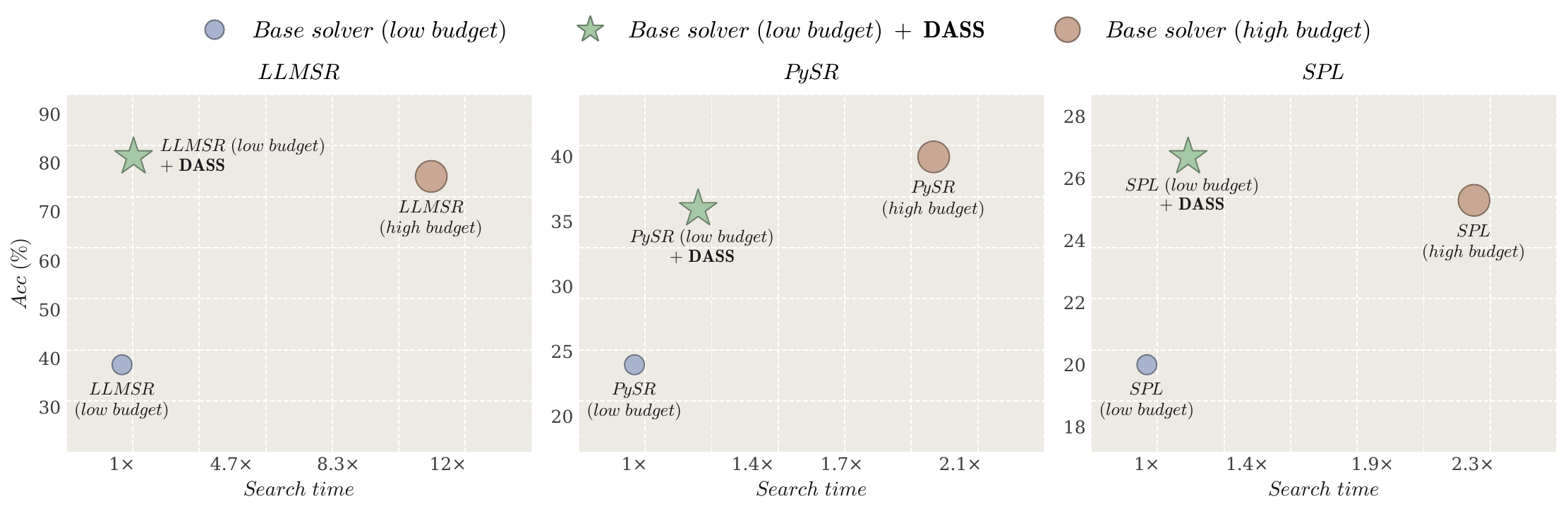}
  \caption{
    Budget efficiency on the Material Science domain; the setup is identical to Fig.~\ref{fig:phys_budget}.
  }
  \label{fig:matsci_budget}
\end{figure}

\begin{figure}[htbp]
  \centering
  \includegraphics[width=\linewidth]{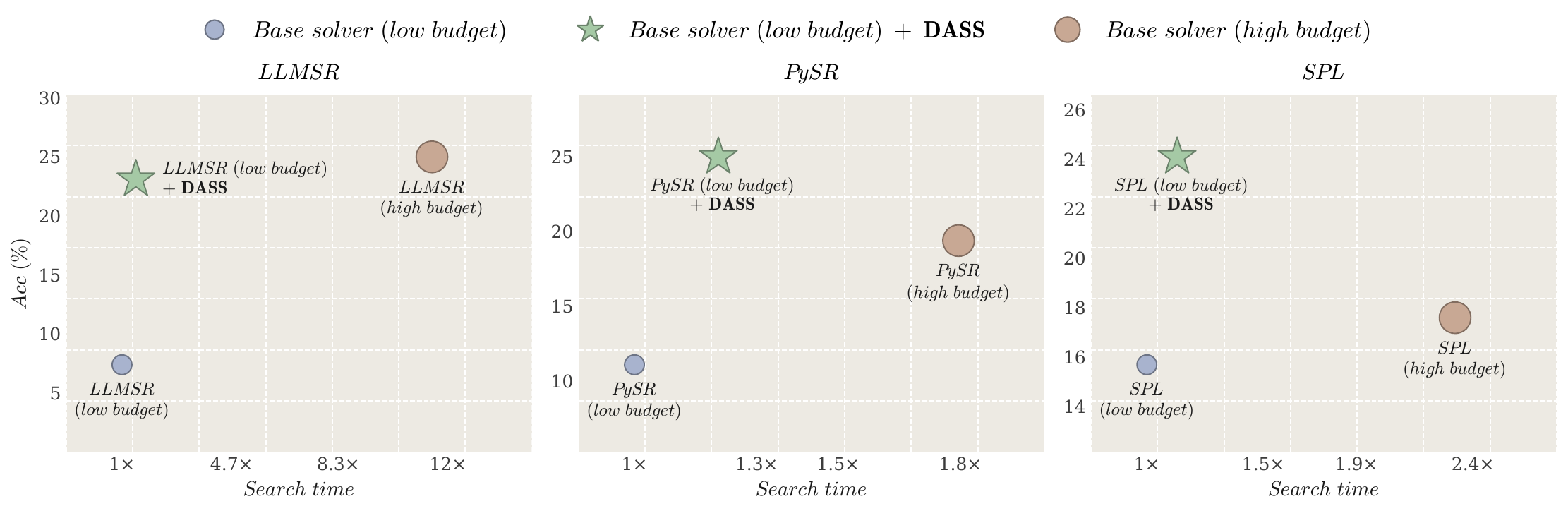}
  \caption{
    Budget efficiency on the Chemistry domain; the setup is identical to Fig.~\ref{fig:phys_budget}.
  }
  \label{fig:chem_budget}
\end{figure}

\begin{figure}[htbp]
  \centering
  \includegraphics[width=\linewidth]{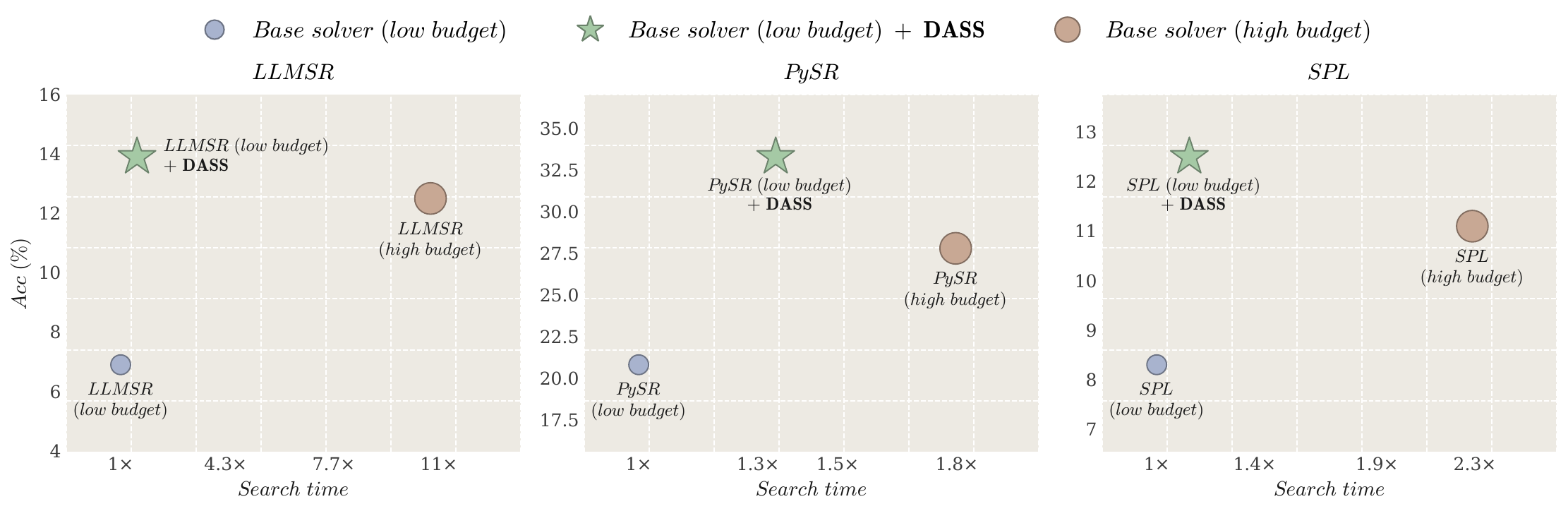}
  \caption{
    Budget efficiency on the Biology domain; the setup is identical to Fig.~\ref{fig:phys_budget}.
  }
  \label{fig:bio_budget}
\end{figure}

Section~\ref{4exp:budget} reports the budget-efficiency results of the three base solvers, LLMSR, PySR, and SPL, averaged over the four scientific domains. Here, we further decompose the averaged results by domain, as shown in Figures~\ref{fig:phys_budget}--\ref{fig:bio_budget}. All domains exhibit a trend consistent with the averaged results: low budget + DASS achieves $\mathrm{Acc}_{0.01}$ values that are comparable to, and in some cases higher than, those of the high-budget base solvers, while substantially reducing search time. This pattern holds consistently across the 12 comparisons formed by three solvers and four domains, suggesting that the guidance benefit provided by DASS under low-budget settings does not depend on solver-specific behavior in any particular domain.

The magnitude of the improvement varies across domains. On Physics and Material Science, increasing the budget of the base solvers from low to high still yields noticeable improvements in $\mathrm{Acc}_{0.01}$, yet low budget + DASS reaches or exceeds the same level of performance. On Chemistry and Biology, the improvement from high budget over low budget is relatively limited, but low budget + DASS still consistently outperforms the high-budget base solvers. Taken together, these results show that, regardless of how sensitive a base solver is to budget expansion, incorporating DASS guidance can provide gains under low-budget settings that are comparable to, or stronger than, those obtained by allocating additional search resources.

This finding is consistent with the stable OOD improvements of DASS reported in Appendix~\ref{app:ood}. When the native search capability of a solver is constrained, either by distribution shift or by limited budget, effective guidance signals can extract more useful information per unit of search budget, thereby substituting for part of the benefit that would otherwise require expanded search resources.

\subsection{Ablation of Subspace Calibration and Reshaping} 
\label{app:abla-of-subspace}

\begin{table}[htbp]
  \centering
  \caption
    {
    Ablation study on the subspace update mechanism, measured by $\mathrm{Acc}_{0.01}$.
    The three settings share the same LLM-generated initial basis term set $\mathcal{B}^{(0)}$ and initial scores $\bm{\omega}^{(0)}$.
    They differ only in which dynamic update mechanisms are enabled after initialization:
    (1) \emph{Frozen prior}, where both $\mathcal{B}$ and $\bm{\omega}$ are frozen after initialization;
    (2) \emph{+ Calibration}, where only subspace calibration is enabled;
    and (3) \emph{+ Reshaping}, where both subspace calibration and subspace reshaping are enabled, so that $\mathcal{B}$ and $\bm{\omega}$ are updated according to search feedback.
    The last setting corresponds to the full DASS.
    }
  \label{tab:closed-loop}
  \resizebox{\linewidth}{!}{\begin{tabular}{l cccc cccc cccc}
\toprule
\multirow{2}[3]{*}{\textbf{Update mechanisms enabled}}
  & \multicolumn{4}{c}{\textbf{LLMSR}}
  & \multicolumn{4}{c}{\textbf{PySR}}
  & \multicolumn{4}{c}{\textbf{SPL}} \\
\cmidrule(lr){2-5}\cmidrule(lr){6-9}\cmidrule(lr){10-13}
 & Phys & MatSci & Chem & Bio
 & Phys & MatSci & Chem & Bio
 & Phys & MatSci & Chem & Bio \\
\midrule
Frozen prior ($\mathcal{B}, \boldsymbol{\omega}$ fixed after init)                              & 17.4 & 64.0 & 13.9 & 9.7  & 20.5          & 29.3          & 18.5          & 26.4          & 14.4          & 22.7          & 19.4          & 11.1 \\
+ Calibration (only $\boldsymbol{\omega}$ updated)                                              & 25.0 & 73.3 & 20.4 & 12.5 & \textbf{22.7} & 33.3          & 23.1          & 30.6          & 19.7          & 24.0          & 22.2          & 11.1 \\
\textbf{+ Reshaping (DASS; $\mathcal{B}, \boldsymbol{\omega}$ both updated)}               & \textbf{29.5} & \textbf{81.3} & \textbf{23.1} & \textbf{13.9} & \textbf{22.7} & \textbf{36.0} & \textbf{25.0} & \textbf{33.3} & \textbf{22.0} & \textbf{26.7} & \textbf{24.1} & \textbf{12.5} \\
\bottomrule
\end{tabular}%
}
\end{table}

Table~\ref{tab:main-results} demonstrates the effectiveness of the complete DASS framework. However, it does not by itself reveal whether the gains should be attributed to the subspace calibration, the subspace reshaping, or merely the high-quality initial prior provided by the LLM. Table~\ref{tab:closed-loop} disentangles these factors. The three settings share the same initialization $(\mathcal{B}^{(0)}, \bm{\omega}^{(0)})$. Enabling calibration alone yields an average relative improvement of $21.33\%$ in $\mathrm{Acc}_{0.01}$ over the Frozen-prior setting. This result indicates that even when $\mathcal{B}$ is fixed, multi-environment evaluation and Gibbs posterior updates can redistribute $\bm{\omega}$ and substantially improve search guidance. After further adding subspace reshaping, the overall performance obtains an additional relative improvement of approximately $10.22\%$ in $\mathrm{Acc}_{0.01}$, which demonstrates that high-quality solutions returned by the solver can further enrich the basis term set. The only tie occurs for PySR on the Physics dataset, where + Calibration and full DASS both achieve $22.7$.

\section{Hyperparameter Sensitivity and Basis Term Set Initialization}
\label{app:sensitivity}

In addition to the algorithmic constants fixed in the main text, namely $\beta$, $\eta$, and $\omega_0$, DASS has four search-stage hyperparameters: the number of samples per round $m$, the basis-set size $n$, the elite ratio $k$, and the number of DASS rounds per task $T$. Sections~\ref{app:sensitivity-m}--\ref{app:sensitivity-T} conduct ablation studies on these four hyperparameters. The base solvers are LLM-SR, PySR, and SPL, and the evaluation metric is the average $\mathrm{Acc}_{0.01}$ over the four domains on the LSR-Synth ID test set. Section~\ref{app:prompt} provides the GPT-4o prompt used to initialize the basis set $\mathcal{B}^{(0)}$.

\subsection{Sensitivity to Sample Count $m$}
\label{app:sensitivity-m}

\begin{figure}[htbp]
  \centering
  \includegraphics[width=\linewidth]{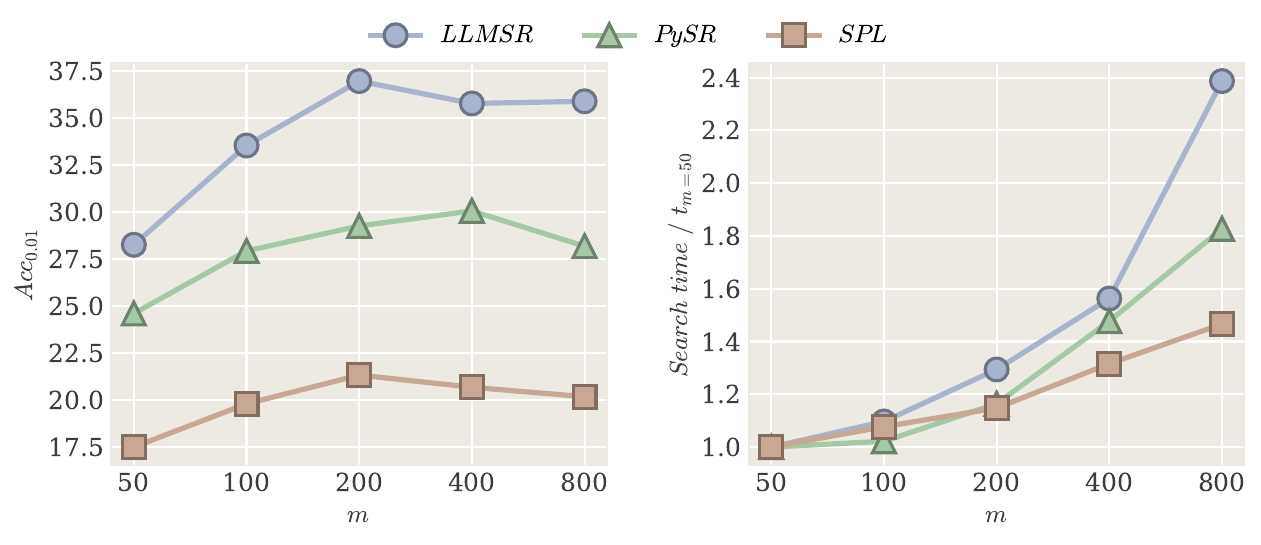}
  \caption{
    Sensitivity to the sample count $m$. Left: four-domain average $\mathrm{Acc}_{0.01}$ plotted against $m$, with the three curves corresponding to LLM-SR, PySR, and SPL. Right: the corresponding average per-problem search time for each value of $m$, measured as a multiple of the search time at $m=50$.
  }
  \label{fig:sensitivity_m}
\end{figure}

As defined in Section~\ref{3method:Calibration}, the sample count $m$ specifies the number of candidate expressions drawn from the current auxiliary subspace in each DASS round. It therefore controls the trade-off between search coverage and the cost of sampling and evaluation. Fig.~\ref{fig:sensitivity_m} sweeps $m\in\{50,100,200,400,800\}$. All three solvers improve markedly as $m$ increases from 50 to 200. Beyond $m=200$, the gains do not continue: from $m=200$ to $m=400$, $\mathrm{Acc}_{0.01}$ decreases slightly for LLM-SR ($36.95$ to $35.76$) and SPL ($21.33$ to $20.68$) and increases slightly for PySR ($29.25$ to $30.05$), while the per-task search time continues to grow. At $m=800$, all three solvers remain below their $m=200$ accuracy. This pattern suggests that, once the useful sampling coverage is reached, additional samples provide limited new evidence for updating the elite set. We therefore use $m=200$ as the default: it gives the highest $\mathrm{Acc}_{0.01}$ for LLM-SR and SPL and is within $0.8$ points of the best value for PySR, at substantially lower search time than $m=400$.

\subsection{Sensitivity to Basis-Set Size $n$}
\label{app:sensitivity-n}

\begin{figure}[htbp]
  \centering
  \includegraphics[width=0.5\linewidth]{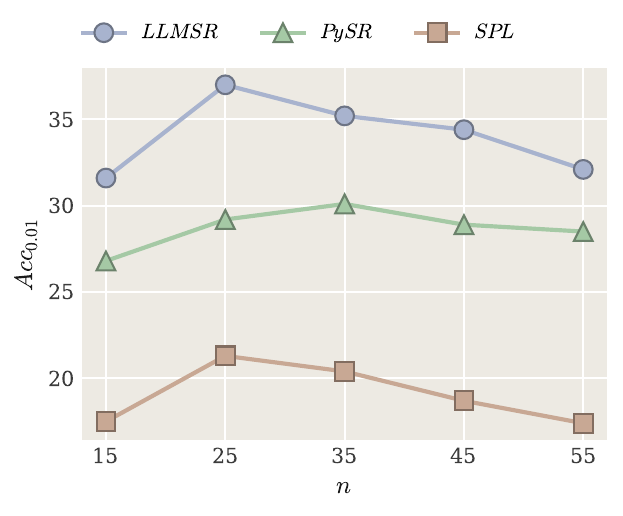}
  \caption{
    Sensitivity to the basis-set size $n$. 4-domain mean $\mathrm{Acc}_{0.01}$ for LLM-SR, PySR, and SPL with $n\in\{15,25,35,45,55\}$. 
  }
  \label{fig:sensitivity_n}
\end{figure}

As defined in Section~\ref{3method:problem}, the basis-set size $n=|\mathcal{B}|$ denotes the number of basis terms retained by DASS, and therefore affects the set of candidate structures considered during active utility tracking. Fig.~\ref{fig:sensitivity_n} sweeps $n\in\{15,25,35,45,55\}$. All three solvers exhibit an inverted-U-shaped trend, with the peak occurring around $n\!\in\![25,35]$, and show clear degradation at $n=55$. A smaller $n$ may underrepresent the auxiliary subspace, whereas a larger $n$ may dilute the potential-score distribution $\bm{\omega}^{(t)}$ and make it more difficult for elite evidence to concentrate on a small number of useful terms. We set the default value to $n=25$, which lies at or near the peak for each solver while keeping the basis set sufficiently small, corresponding to a lower-dimensional auxiliary subspace.

\subsection{Sensitivity to Elite Ratio $k$}
\label{app:sensitivity-k}

\begin{figure}[htbp]
  \centering
  \includegraphics[width=0.5\linewidth]{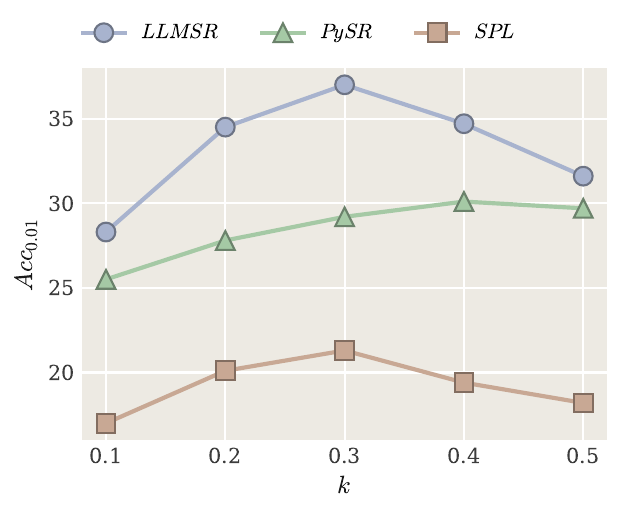}
  \caption{
    Sensitivity to the elite ratio $k$. 4-domain mean $\mathrm{Acc}_{0.01}$ for LLM-SR, PySR, and SPL with $k\in\{10\%,20\%,30\%,40\%,50\%\}$.
  }
  \label{fig:sensitivity_k}
\end{figure}

The elite ratio $k$ (in percentage), introduced in Section~\ref{3method:Calibration}, selects the top-$k\%$ candidates in $\mathcal{Q}^{(t)}$ for positive utility assignment to basis terms. Fig.~\ref{fig:sensitivity_k} reports the sweep over $k\in\{10\%,20\%,30\%,40\%,50\%\}$. The curves increase from $k=10\%$; LLM-SR and SPL reach their peaks at $k=30\%$, while PySR peaks at $k=40\%$. The peak is relatively flat for PySR but more pronounced for LLM-SR. The overall trend across $k$ suggests that a very small elite slice may provide a utility signal that is too selective and therefore less stable, whereas a very large slice may include expressions with weaker fitness and reduce the contrast between useful and less informative terms. We set the default to $k=30\%$, which lies in the favorable range for all three solvers.

\subsection{Sensitivity to Insertion Count $T$}
\label{app:sensitivity-T}

\begin{figure}[htbp]
  \centering
  \includegraphics[width=\linewidth]{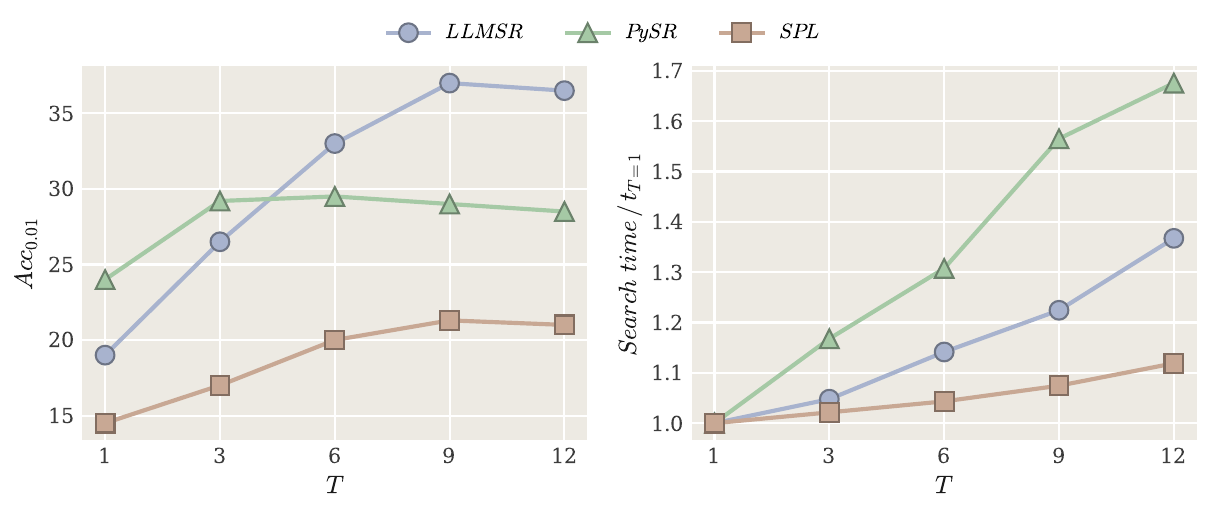}
  \caption{
    Sensitivity to the insertion count $T$. Left: 4-domain mean $\mathrm{Acc}_{0.01}$ for LLM-SR, PySR, and SPL with $T\in\{1,3,6,9,12\}$. Right: the corresponding average per-problem search time for each value of $T$, measured as a multiple of the search time at $T=1$.
  }
  \label{fig:sensitivity_T}
\end{figure}

DASS is invoked $T$ times within the overall search budget of the base solver. Fig.~\ref{fig:sensitivity_T} reports the sweep over $T\in\{1,3,6,9,12\}$. The $\mathrm{Acc}_{0.01}$ of all three solvers improves noticeably from $T=1$ to $T=3$, but their trends diverge thereafter: PySR largely saturates around $T=3$, with $T=6$ bringing only about 0.3 absolute percentage points of marginal improvement, and slightly regresses as $T$ is further increased; LLM-SR and SPL continue to benefit until $T=9$, after which their performance levels off. The search time increases approximately linearly with $T$.

This trend indicates that increasing $T$ within a certain range can refresh the auxiliary subspace more frequently and provide the base solver with more opportunities to absorb structural feedback. However, once the refresh frequency exceeds the range that the solver can effectively exploit, the benefit of additional invocations diminishes quickly, while the computational overhead continues to grow approximately linearly. The earlier saturation of PySR may be related to the relatively high evolutionary efficiency of its GP search: after a small number of refreshes, the injected subspace may already be sufficiently exploited, whereas overly frequent insertions may interfere with the continuity of evolution. In contrast, LLM-SR and SPL may be more sensitive to continued refreshes of structural guidance, and therefore still obtain gains at larger values of $T$. Considering both accuracy and cost, we set the default to $T=3$ for PySR and $T=9$ for LLM-SR and SPL.

\subsection{Prompt for Initializing the Basis Term Set $\mathcal{B}$}
\label{app:prompt}

\begin{figure}[htbp]
  \centering
  \includegraphics[width=\linewidth]{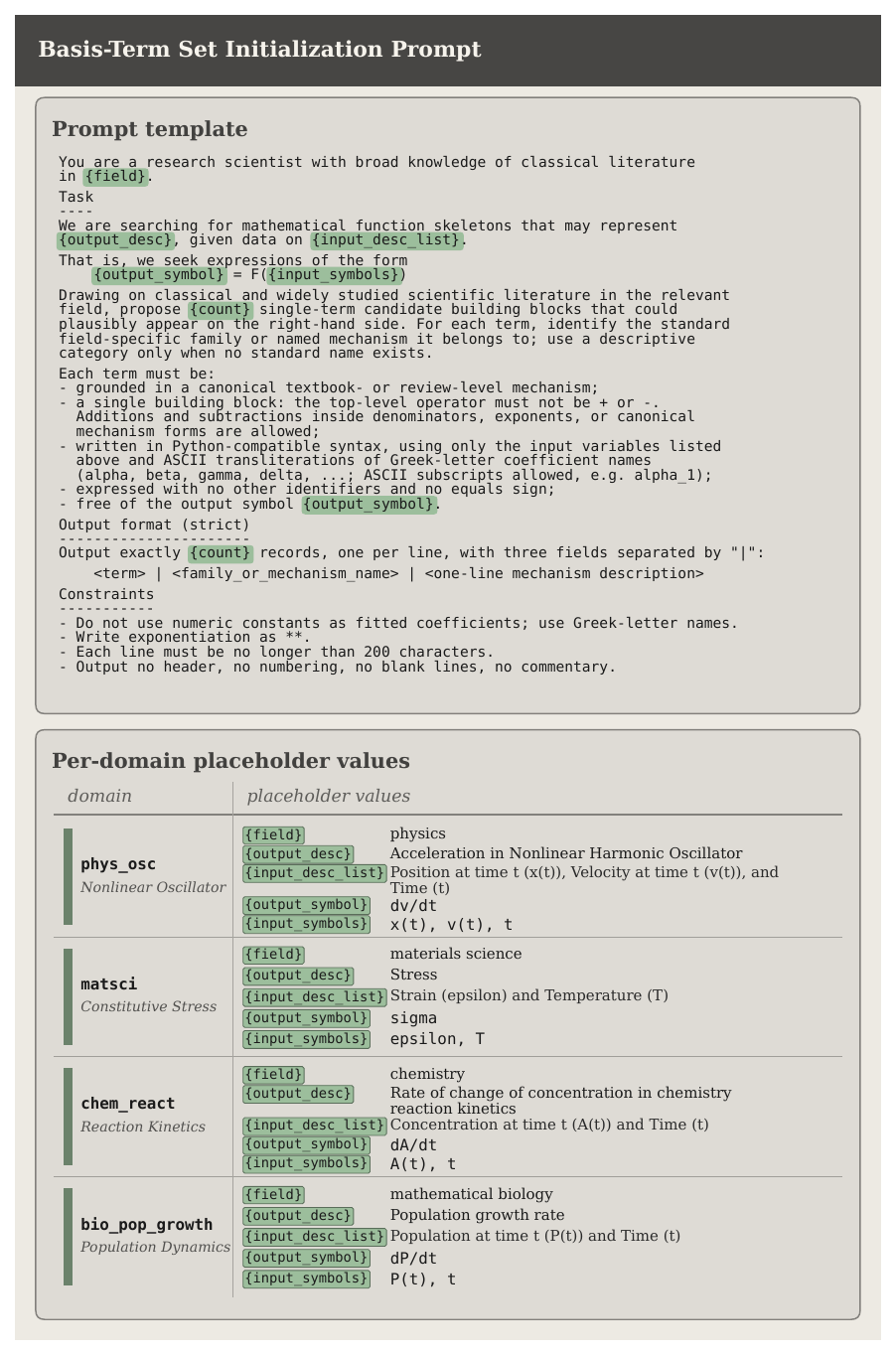}
  \caption{
    Overview of the prompt used to initialize the basis term set $\mathcal{B}$. Top: a unified prompt template with six placeholders, including five domain-related fields and \texttt{\{count\}}, which specifies the number of terms to generate. Bottom: domain-placeholder values for Physics, Materials Science, Chemistry, and Biology in LLM-SRBench, instantiated from benchmark-provided metadata available as task input, not from target equations or evaluation data.
  }
  \label{fig:prompt}
\end{figure}

We instantiate a unified prompt template for each domain (Fig.~\ref{fig:prompt}) and use GPT-4o to initialize the basis term set $\mathcal{B}$, resulting in the initial set $\mathcal{B}^{(0)}$. GPT-4o is used to remain consistent with the LLM employed by LLM-SRBench~\citep{llmsrbench} during its synthesis stage. The template contains five main domain-related placeholder fields: the domain, textual descriptions of the output and input variables, and the output and input symbols. These fields are filled according to the description of each scientific domain.
The \texttt{\{count\}} field is set to 10, so that each call asks the model to return 10 candidate terms. Under memory-free sampling, the same prompt is independently queried 20 times (with sampling temperature drawn independently from $U(0.7, 1.1)$ per call), yielding 200 candidate records for each domain. Before computing frequencies, we normalize the names of all free coefficients in each record by replacing them with a shared placeholder. Records that differ only in the naming of coefficient symbols are therefore merged into the same unique term; for example, $\alpha\sin(x)$ and $\beta\sin(x)$ are treated as equivalent after merging. Let $c(s_i)$ denote the frequency of the merged term $s_i$. We select the top-$n$ unique terms ranked by frequency to form the initial basis set $\mathcal{B}^{(0)}$ for the corresponding domain, where $n$ is the upper bound on the basis-set size. The normalized frequency is used as the initial potential score, $\omega^{(0)}_i=\frac{c(s_i)}{\sum^n_{j=1}c(s_j)}$. Figures~\ref{fig:prompt_example_phys_osc}, \ref{fig:prompt_example_matsci}, \ref{fig:prompt_example_chem_react}, and \ref{fig:prompt_example_bio_pop_growth} present representative prompt input--output examples for four datasets across distinct scientific domains: Physics, Materials Science, Chemistry, and Biology, respectively.

\definecolor{boxfill}{rgb}{0.95,0.97,1.0}
\definecolor{boxstroke}{rgb}{0.2,0.4,0.7}

\newsavebox{\promptBodyPhysOsc}
\begin{lrbox}{\promptBodyPhysOsc}
\begin{minipage}{13cm}
\begin{Verbatim}[fontsize=\tiny]
You are a research scientist with broad knowledge of classical literature
in physics.

Task
----
We are searching for mathematical function skeletons that may represent
Acceleration in Nonlinear Harmonic Oscillator, given data on Position at time t (x(t)), Velocity at
time t (v(t)), and Time (t).

That is, we seek expressions of the form
    dv/dt = F(x(t), v(t), t)

Drawing on classical and widely studied scientific literature in the relevant
field, propose 10 single-term candidate building blocks that could
plausibly appear on the right-hand side. For each term, identify the standard
field-specific family or named mechanism it belongs to; use a descriptive
category only when no standard name exists.

Each term must be:
- grounded in a canonical textbook- or review-level mechanism;
- a single building block: the top-level operator must not be + or -.
  Additions and subtractions inside denominators, exponents, or canonical
  mechanism forms are allowed;
- written in Python-compatible syntax, using only the input variables listed
  above and ASCII transliterations of Greek-letter coefficient names
  (alpha, beta, gamma, delta, ...; ASCII subscripts allowed, e.g. alpha_1);
- expressed with no other identifiers and no equals sign;
- free of the output symbol dv/dt.

Output format (strict)
----------------------
Output exactly 10 records, one per line, with three fields separated by "|":

    <term> | <family_or_mechanism_name> | <one-line mechanism description>

Constraints
-----------
- Do not use numeric constants as fitted coefficients; use Greek-letter names.
- Write exponentiation as **.
- Each line must be no longer than 200 characters.
- Output no header, no numbering, no blank lines, no commentary.
\end{Verbatim}
\end{minipage}
\end{lrbox}

\begin{figure}[!t]
\centering
\begin{tikzpicture}
    \node[
        draw=boxstroke,
        line width=1pt,
        rounded corners=10pt,
        fill=boxfill,
        inner sep=5pt,
        align=left,
        text width=13cm
    ] {%
        {\tiny\textbf{Prompt}}\par\vspace{2pt}
        \usebox{\promptBodyPhysOsc}%
    };
\end{tikzpicture}
\par\vspace{4pt}
\begin{tikzpicture}
    \node[
        draw=boxstroke,
        line width=1pt,
        rounded corners=10pt,
        fill=boxfill,
        inner sep=5pt,
        align=left,
        text width=13cm
    ] {%
        {\tiny\textbf{Output}}\par\vspace{2pt}
        {\tiny
        \setlength{\tabcolsep}{5pt}%
        \renewcommand{\arraystretch}{1.2}%
        \begin{tabular}{@{}l@{\,\textbar\,}l@{\,\textbar\,}l@{}}
\texttt{alpha*x} & Linear Damping & Velocity-dependent linear friction force proportional to speed \\
\texttt{gamma*x} & Linear Spring Force & Restoring force proportional to displacement from equilibrium \\
\texttt{beta*v**2} & Quadratic Drag & Air resistance proportional to the square of velocity \\
\texttt{delta*v} & Linear Viscosity & Newtonian viscous damping coefficient times velocity \\
\texttt{epsilon*x**3} & Duffing Nonlinearity & Cubic stiffness term in hardening or softening springs \\
\texttt{zeta*x*v} & Gyroscopic Coupling & Velocity-dependent force linear in both position and speed \\
\texttt{phi*x*sin(gamma*t)} & Time-Modulated Spring & External periodic forcing with time-varying amplitude \\
\texttt{psi*v*cos(gamma*t)} & Time-Modulated Drag & Velocity-dependent resistance with time-varying coefficient \\
\texttt{chi*x**2} & Quadratic Spring & Nonlinear restoring force scaling with the square of displacement \\
\texttt{sigma*exp(-tau*x)} & Saturating Restitution & Exponentially decaying force representing saturation effects \\
        \end{tabular}%
        }%
    };
\end{tikzpicture}

\caption{Prompt used to generate the initial basis terms set
$\mathcal{B}^{(0)}$ for the Physics subset of LLM-SRBench, with the
benchmark-revealed placeholder values substituted in (top), and the sample LLM
output it elicits in a single call (bottom). Each output record carries three
fields: \texttt{term | family/mechanism | one-line description}. Ten such
records are requested via \texttt{\{count\}}\,=\,10. LLM backbone: GPT-4o.}
\label{fig:prompt_example_phys_osc}
\end{figure}

\definecolor{boxfill}{rgb}{0.95,0.97,1.0}
\definecolor{boxstroke}{rgb}{0.2,0.4,0.7}

\newsavebox{\promptBodyMatsci}
\begin{lrbox}{\promptBodyMatsci}
\begin{minipage}{13cm}
\begin{Verbatim}[fontsize=\tiny]
You are a research scientist with broad knowledge of classical literature
in materials science.

Task
----
We are searching for mathematical function skeletons that may represent
Stress, given data on Strain (epsilon) and Temperature (T).

That is, we seek expressions of the form
    sigma = F(epsilon, T)

Drawing on classical and widely studied scientific literature in the relevant
field, propose 10 single-term candidate building blocks that could
plausibly appear on the right-hand side. For each term, identify the standard
field-specific family or named mechanism it belongs to; use a descriptive
category only when no standard name exists.

Each term must be:
- grounded in a canonical textbook- or review-level mechanism;
- a single building block: the top-level operator must not be + or -.
  Additions and subtractions inside denominators, exponents, or canonical
  mechanism forms are allowed;
- written in Python-compatible syntax, using only the input variables listed
  above and ASCII transliterations of Greek-letter coefficient names
  (alpha, beta, gamma, delta, ...; ASCII subscripts allowed, e.g. alpha_1);
- expressed with no other identifiers and no equals sign;
- free of the output symbol sigma.

Output format (strict)
----------------------
Output exactly 10 records, one per line, with three fields separated by "|":

    <term> | <family_or_mechanism_name> | <one-line mechanism description>

Constraints
-----------
- Do not use numeric constants as fitted coefficients; use Greek-letter names.
- Write exponentiation as **.
- Each line must be no longer than 200 characters.
- Output no header, no numbering, no blank lines, no commentary.
\end{Verbatim}
\end{minipage}
\end{lrbox}

\begin{figure}[!t]
\centering
\begin{tikzpicture}
    \node[
        draw=boxstroke,
        line width=1pt,
        rounded corners=10pt,
        fill=boxfill,
        inner sep=5pt,
        align=left,
        text width=13cm
    ] {%
        {\tiny\textbf{Prompt}}\par\vspace{2pt}
        \usebox{\promptBodyMatsci}%
    };
\end{tikzpicture}
\par\vspace{4pt}
\begin{tikzpicture}
    \node[
        draw=boxstroke,
        line width=1pt,
        rounded corners=10pt,
        fill=boxfill,
        inner sep=5pt,
        align=left,
        text width=13cm
    ] {%
        {\tiny\textbf{Output}}\par\vspace{2pt}
        {\tiny
        \setlength{\tabcolsep}{5pt}%
        \renewcommand{\arraystretch}{1.2}%
        \begin{tabular}{@{}l@{\,\textbar\,}l@{\,\textbar\,}l@{}}
\texttt{alpha*epsilon} & Linear Elasticity & Hooke's law stress proportional to strain \\
\texttt{beta*epsilon*gamma} & Thermomechanical Coupling & Linear thermoelastic expansion stress \\
\texttt{gamma/(1+delta*epsilon)} & Nonlinear Elasticity & Polynomial series expansion term \\
\texttt{alpha*epsilon/(1+beta*T)} & Thermal Softening & Temperature-dependent modulus reduction \\
\texttt{beta*epsilon**2} & Geometric Nonlinearity & Second-order strain hardening term \\
\texttt{gamma*exp(-delta*T)*epsilon} & Arrhenius Thermoelasticity & Thermally activated barrier reduction \\
\texttt{alpha*epsilon/(1-delta*epsilon)} & Plastic Flow & Denominator-based saturation limit \\
\texttt{beta*epsilon/(1+gamma*T)} & Thermal Expansion Stress & Constraint-induced thermal stress term \\
\texttt{gamma*epsilon/(1+beta*T+delta*epsilon**2)} & Viscoelastic Relaxation & Combined thermal and strain rate effects \\
\texttt{alpha*epsilon*exp(-beta/T)} & Thermally Activated Flow & Arrhenius rate-dependent stress contribution \\
        \end{tabular}%
        }%
    };
\end{tikzpicture}

\caption{Prompt used to generate the initial basis terms set
$\mathcal{B}^{(0)}$ for the Material Science subset of LLM-SRBench, with the
benchmark-revealed placeholder values substituted in (top), and the sample LLM
output it elicits in a single call (bottom). Each output record carries three
fields: \texttt{term | family/mechanism | one-line description}. Ten such
records are requested via \texttt{\{count\}}\,=\,10. LLM backbone: GPT-4o.}
\label{fig:prompt_example_matsci}
\end{figure}

\definecolor{boxfill}{rgb}{0.95,0.97,1.0}
\definecolor{boxstroke}{rgb}{0.2,0.4,0.7}

\newsavebox{\promptBodyChemReact}
\begin{lrbox}{\promptBodyChemReact}
\begin{minipage}{13cm}
\begin{Verbatim}[fontsize=\tiny]
You are a research scientist with broad knowledge of classical literature
in chemistry.

Task
----
We are searching for mathematical function skeletons that may represent
Rate of change of concentration in chemistry reaction kinetics, given data on Concentration at time
t (A(t)) and Time (t).

That is, we seek expressions of the form
    dA/dt = F(A(t), t)

Drawing on classical and widely studied scientific literature in the relevant
field, propose 10 single-term candidate building blocks that could
plausibly appear on the right-hand side. For each term, identify the standard
field-specific family or named mechanism it belongs to; use a descriptive
category only when no standard name exists.

Each term must be:
- grounded in a canonical textbook- or review-level mechanism;
- a single building block: the top-level operator must not be + or -.
  Additions and subtractions inside denominators, exponents, or canonical
  mechanism forms are allowed;
- written in Python-compatible syntax, using only the input variables listed
  above and ASCII transliterations of Greek-letter coefficient names
  (alpha, beta, gamma, delta, ...; ASCII subscripts allowed, e.g. alpha_1);
- expressed with no other identifiers and no equals sign;
- free of the output symbol dA/dt.

Output format (strict)
----------------------
Output exactly 10 records, one per line, with three fields separated by "|":

    <term> | <family_or_mechanism_name> | <one-line mechanism description>

Constraints
-----------
- Do not use numeric constants as fitted coefficients; use Greek-letter names.
- Write exponentiation as **.
- Each line must be no longer than 200 characters.
- Output no header, no numbering, no blank lines, no commentary.
\end{Verbatim}
\end{minipage}
\end{lrbox}

\begin{figure}[!t]
\centering
\begin{tikzpicture}
    \node[
        draw=boxstroke,
        line width=1pt,
        rounded corners=10pt,
        fill=boxfill,
        inner sep=5pt,
        align=left,
        text width=13cm
    ] {%
        {\tiny\textbf{Prompt}}\par\vspace{2pt}
        \usebox{\promptBodyChemReact}%
    };
\end{tikzpicture}
\par\vspace{4pt}
\begin{tikzpicture}
    \node[
        draw=boxstroke,
        line width=1pt,
        rounded corners=10pt,
        fill=boxfill,
        inner sep=5pt,
        align=left,
        text width=13cm
    ] {%
        {\tiny\textbf{Output}}\par\vspace{2pt}
        {\tiny
        \setlength{\tabcolsep}{5pt}%
        \renewcommand{\arraystretch}{1.2}%
        \begin{tabular}{@{}l@{\,\textbar\,}l@{\,\textbar\,}l@{}}
\texttt{alpha*A(t)*t**-1} & First-Order Decay & Rate proportional to concentration and inverse time scaling \\
\texttt{beta*A(t)**2} & Second-Order Bimolecular & Rate proportional to square of reactant concentration \\
\texttt{gamma*A(t)**0.5} & Half-Order Mechanism & Rate proportional to square root of concentration \\
\texttt{delta*A(t)*t**2} & Autocatalytic Growth & Rate proportional to concentration and squared time \\
\texttt{epsilon*A(t)*exp(-A(t))} & Surface-Site Saturation & Rate limited by available surface sites relative to coverage \\
\texttt{zeta*A(t)**2*A(t)**-1} & Effective First-Order & Simplified form of second-order with competing terms \\
\texttt{eta*A(t)*sin(t)} & Oscillatory Reaction & Rate modulated by periodic temporal factors \\
\texttt{theta*A(t)**-1} & Inverse Concentration & Rate inversely proportional to concentration \\
\texttt{iota*A(t)*t**0.5} & Diffusion-Limited Growth & Rate proportional to concentration and square root of time \\
\texttt{kappa*A(t)*log(A(t))} & Logarithmic Inhibition & Rate suppressed logarithmically by concentration \\
        \end{tabular}%
        }%
    };
\end{tikzpicture}

\caption{Prompt used to generate the initial basis terms set
$\mathcal{B}^{(0)}$ for the Chemistry subset of LLM-SRBench, with the
benchmark-revealed placeholder values substituted in (top), and the sample LLM
output it elicits in a single call (bottom). Each output record carries three
fields: \texttt{term | family/mechanism | one-line description}. Ten such
records are requested via \texttt{\{count\}}\,=\,10. LLM backbone: GPT-4o.}
\label{fig:prompt_example_chem_react}
\end{figure}

\definecolor{boxfill}{rgb}{0.95,0.97,1.0}
\definecolor{boxstroke}{rgb}{0.2,0.4,0.7}

\newsavebox{\promptBodyBioPopGrowth}
\begin{lrbox}{\promptBodyBioPopGrowth}
\begin{minipage}{13cm}
\begin{Verbatim}[fontsize=\tiny]
You are a research scientist with broad knowledge of classical literature
in mathematical biology.

Task
----
We are searching for mathematical function skeletons that may represent
Population growth rate, given data on Population at time t (P(t)) and Time (t).

That is, we seek expressions of the form
    dP/dt = F(P(t), t)

Drawing on classical and widely studied scientific literature in the relevant
field, propose 10 single-term candidate building blocks that could
plausibly appear on the right-hand side. For each term, identify the standard
field-specific family or named mechanism it belongs to; use a descriptive
category only when no standard name exists.

Each term must be:
- grounded in a canonical textbook- or review-level mechanism;
- a single building block: the top-level operator must not be + or -.
  Additions and subtractions inside denominators, exponents, or canonical
  mechanism forms are allowed;
- written in Python-compatible syntax, using only the input variables listed
  above and ASCII transliterations of Greek-letter coefficient names
  (alpha, beta, gamma, delta, ...; ASCII subscripts allowed, e.g. alpha_1);
- expressed with no other identifiers and no equals sign;
- free of the output symbol dP/dt.

Output format (strict)
----------------------
Output exactly 10 records, one per line, with three fields separated by "|":

    <term> | <family_or_mechanism_name> | <one-line mechanism description>

Constraints
-----------
- Do not use numeric constants as fitted coefficients; use Greek-letter names.
- Write exponentiation as **.
- Each line must be no longer than 200 characters.
- Output no header, no numbering, no blank lines, no commentary.
\end{Verbatim}
\end{minipage}
\end{lrbox}

\begin{figure}[!t]
\centering
\begin{tikzpicture}
    \node[
        draw=boxstroke,
        line width=1pt,
        rounded corners=10pt,
        fill=boxfill,
        inner sep=5pt,
        align=left,
        text width=13cm
    ] {%
        {\tiny\textbf{Prompt}}\par\vspace{2pt}
        \usebox{\promptBodyBioPopGrowth}%
    };
\end{tikzpicture}
\par\vspace{4pt}
\begin{tikzpicture}
    \node[
        draw=boxstroke,
        line width=1pt,
        rounded corners=10pt,
        fill=boxfill,
        inner sep=5pt,
        align=left,
        text width=13cm
    ] {%
        {\tiny\textbf{Output}}\par\vspace{2pt}
        {\tiny
        \setlength{\tabcolsep}{5pt}%
        \renewcommand{\arraystretch}{1.2}%
        \begin{tabular}{@{}l@{\,\textbar\,}l@{\,\textbar\,}l@{}}
\texttt{alpha*P} & Linear Growth & Exponential increase proportional to current population size \\
\texttt{beta*P*(1-P/K)} & Logistic Growth & Density-dependent regulation reducing growth as population approaches carrying capacity \\
\texttt{alpha*P**0.5} & Allee Effect & Reduced per capita growth at low population densities due to mate limitation \\
\texttt{gamma*P**1.5} & Superlinear Growth & Accelerating growth often seen in early epidemic spread or facilitated diffusion \\
\texttt{delta/(1+beta*P)} & Type II Functional Response & Saturation in per capita growth rate as density increases \\
\texttt{epsilon*P*(1-P)**2} & Strong Allee Effect & Critical threshold where growth becomes negative below a specific density \\
\texttt{zeta*P*exp(-beta*P)} & Ricker Model & Discrete-time dynamics with overcompensatory density dependence \\
\texttt{eta*P*sin(alpha*t)} & Seasonal Forcing & Time-dependent modulation of growth rates due to environmental cycles \\
\texttt{theta*P/(1+beta*t)} & Declining Resources & Growth limited by time-dependent depletion of renewable resources \\
\texttt{iota*P**0.8} & Sublinear Growth & Decelerating growth typical of limited dispersal or spatial constraints \\
        \end{tabular}%
        }%
    };
\end{tikzpicture}

\caption{Prompt used to generate the initial basis terms set
$\mathcal{B}^{(0)}$ for the Biology subset of LLM-SRBench, with the
benchmark-revealed placeholder values substituted in (top), and the sample LLM
output it elicits in a single call (bottom). Each output record carries three
fields: \texttt{term | family/mechanism | one-line description}. Ten such
records are requested via \texttt{\{count\}}\,=\,10. LLM backbone: GPT-4o.}
\label{fig:prompt_example_bio_pop_growth}
\end{figure}

\clearpage
\section*{NeurIPS Paper Checklist}

\begin{enumerate}

\item {\bf Claims}
    \item[] Question: Do the main claims made in the abstract and introduction accurately reflect the paper's contributions and scope?
    \item[] Answer: \answerYes{} %
    \item[] Justification: {The abstract and Section~\ref{sec:intro} enumerate three contributions (a new direction, the DASS framework, and a closed-loop update mechanism), each of which is realized in Section~\ref{sec:method} and validated in Section~\ref{sec:exp} and Appendices~\ref{app:additional-results}–\ref{app:sensitivity}.}
    \item[] Guidelines:
    \begin{itemize}
        \item The answer \answerNA{} means that the abstract and introduction do not include the claims made in the paper.
        \item The abstract and/or introduction should clearly state the claims made, including the contributions made in the paper and important assumptions and limitations. A \answerNo{} or \answerNA{} answer to this question will not be perceived well by the reviewers. 
        \item The claims made should match theoretical and experimental results, and reflect how much the results can be expected to generalize to other settings. 
        \item It is fine to include aspirational goals as motivation as long as it is clear that these goals are not attained by the paper. 
    \end{itemize}

\item {\bf Limitations}
    \item[] Question: Does the paper discuss the limitations of the work performed by the authors?
    \item[] Answer: \answerYes{} %
    \item[] Justification: Section~\ref{sec:conclusion} explicitly discusses the dependence of DASS on the downstream solver's guidance-injection interface.
    \item[] Guidelines:
    \begin{itemize}
        \item The answer \answerNA{} means that the paper has no limitation while the answer \answerNo{} means that the paper has limitations, but those are not discussed in the paper. 
        \item The authors are encouraged to create a separate ``Limitations'' section in their paper.
        \item The paper should point out any strong assumptions and how robust the results are to violations of these assumptions (e.g., independence assumptions, noiseless settings, model well-specification, asymptotic approximations only holding locally). The authors should reflect on how these assumptions might be violated in practice and what the implications would be.
        \item The authors should reflect on the scope of the claims made, e.g., if the approach was only tested on a few datasets or with a few runs. In general, empirical results often depend on implicit assumptions, which should be articulated.
        \item The authors should reflect on the factors that influence the performance of the approach. For example, a facial recognition algorithm may perform poorly when image resolution is low or images are taken in low lighting. Or a speech-to-text system might not be used reliably to provide closed captions for online lectures because it fails to handle technical jargon.
        \item The authors should discuss the computational efficiency of the proposed algorithms and how they scale with dataset size.
        \item If applicable, the authors should discuss possible limitations of their approach to address problems of privacy and fairness.
        \item While the authors might fear that complete honesty about limitations might be used by reviewers as grounds for rejection, a worse outcome might be that reviewers discover limitations that aren't acknowledged in the paper. The authors should use their best judgment and recognize that individual actions in favor of transparency play an important role in developing norms that preserve the integrity of the community. Reviewers will be specifically instructed to not penalize honesty concerning limitations.
    \end{itemize}

\item {\bf Theory assumptions and proofs}
    \item[] Question: For each theoretical result, does the paper provide the full set of assumptions and a complete (and correct) proof?
    \item[] Answer: \answerNA{} %
    \item[] Justification: The paper does not include formal theorems or proofs.
    \item[] Guidelines:
    \begin{itemize}
        \item The answer \answerNA{} means that the paper does not include theoretical results. 
        \item All the theorems, formulas, and proofs in the paper should be numbered and cross-referenced.
        \item All assumptions should be clearly stated or referenced in the statement of any theorems.
        \item The proofs can either appear in the main paper or the supplemental material, but if they appear in the supplemental material, the authors are encouraged to provide a short proof sketch to provide intuition. 
        \item Inversely, any informal proof provided in the core of the paper should be complemented by formal proofs provided in appendix or supplemental material.
        \item Theorems and Lemmas that the proof relies upon should be properly referenced. 
    \end{itemize}

    \item {\bf Experimental result reproducibility}
    \item[] Question: Does the paper fully disclose all the information needed to reproduce the main experimental results of the paper to the extent that it affects the main claims and/or conclusions of the paper (regardless of whether the code and data are provided or not)?
    \item[] Answer: \answerYes{} %
    \item[] Justification: Section~\ref{sec:exp} and Appendices~\ref{app:dataset}, \ref{app:base-solver-configs}, \ref{app:llm-backbones}, \ref{app:dass-integration}, \ref{app:sensitivity}, and \ref{app:prompt} provide dataset details, per-solver hyperparameters, LLM backbones and decoding settings, integration details for each base solver, default values and sweeps for DASS hyperparameters, and the GPT-4o prompt used to initialize the basis term set.
    \item[] Guidelines:
    \begin{itemize}
        \item The answer \answerNA{} means that the paper does not include experiments.
        \item If the paper includes experiments, a \answerNo{} answer to this question will not be perceived well by the reviewers: Making the paper reproducible is important, regardless of whether the code and data are provided or not.
        \item If the contribution is a dataset and\slash or model, the authors should describe the steps taken to make their results reproducible or verifiable. 
        \item Depending on the contribution, reproducibility can be accomplished in various ways. For example, if the contribution is a novel architecture, describing the architecture fully might suffice, or if the contribution is a specific model and empirical evaluation, it may be necessary to either make it possible for others to replicate the model with the same dataset, or provide access to the model. In general. releasing code and data is often one good way to accomplish this, but reproducibility can also be provided via detailed instructions for how to replicate the results, access to a hosted model (e.g., in the case of a large language model), releasing of a model checkpoint, or other means that are appropriate to the research performed.
        \item While NeurIPS does not require releasing code, the conference does require all submissions to provide some reasonable avenue for reproducibility, which may depend on the nature of the contribution. For example
        \begin{enumerate}
            \item If the contribution is primarily a new algorithm, the paper should make it clear how to reproduce that algorithm.
            \item If the contribution is primarily a new model architecture, the paper should describe the architecture clearly and fully.
            \item If the contribution is a new model (e.g., a large language model), then there should either be a way to access this model for reproducing the results or a way to reproduce the model (e.g., with an open-source dataset or instructions for how to construct the dataset).
            \item We recognize that reproducibility may be tricky in some cases, in which case authors are welcome to describe the particular way they provide for reproducibility. In the case of closed-source models, it may be that access to the model is limited in some way (e.g., to registered users), but it should be possible for other researchers to have some path to reproducing or verifying the results.
        \end{enumerate}
    \end{itemize}

\item {\bf Open access to data and code}
    \item[] Question: Does the paper provide open access to the data and code, with sufficient instructions to faithfully reproduce the main experimental results, as described in supplemental material?
    \item[] Answer: \answerYes{} %
    \item[] Justification: We provide a link to our open-source code repository on the first page of the paper: \url{https://github.com/SR-Store/NeurIPS2026-DASS}. The LSR-Synth dataset is publicly available from LLM-SRBench~\citep{llmsrbench}.
    \item[] Guidelines:
    \begin{itemize}
        \item The answer \answerNA{} means that paper does not include experiments requiring code.
        \item Please see the NeurIPS code and data submission guidelines (\url{https://neurips.cc/public/guides/CodeSubmissionPolicy}) for more details.
        \item While we encourage the release of code and data, we understand that this might not be possible, so \answerNo{} is an acceptable answer. Papers cannot be rejected simply for not including code, unless this is central to the contribution (e.g., for a new open-source benchmark).
        \item The instructions should contain the exact command and environment needed to run to reproduce the results. See the NeurIPS code and data submission guidelines (\url{https://neurips.cc/public/guides/CodeSubmissionPolicy}) for more details.
        \item The authors should provide instructions on data access and preparation, including how to access the raw data, preprocessed data, intermediate data, and generated data, etc.
        \item The authors should provide scripts to reproduce all experimental results for the new proposed method and baselines. If only a subset of experiments are reproducible, they should state which ones are omitted from the script and why.
        \item At submission time, to preserve anonymity, the authors should release anonymized versions (if applicable).
        \item Providing as much information as possible in supplemental material (appended to the paper) is recommended, but including URLs to data and code is permitted.
    \end{itemize}

\item {\bf Experimental setting/details}
    \item[] Question: Does the paper specify all the training and test details (e.g., data splits, hyperparameters, how they were chosen, type of optimizer) necessary to understand the results?
    \item[] Answer: \answerYes{} %
    \item[] Justification: Experimental settings---dataset splits ($4{,}000$ train / $500$ ID test / $500$ OOD test per task, see Appendix~\ref{app:dataset}), metrics (NMSE and $\mathrm{Acc}_{0.01}$), search budgets per solver, DASS hyperparameters, and LLM decoding parameters---are specified in Section~\ref{sec:exp} and Appendices~\ref{app:base-solver-configs}--\ref{app:sensitivity}.
    \item[] Guidelines:
    \begin{itemize}
        \item The answer \answerNA{} means that the paper does not include experiments.
        \item The experimental setting should be presented in the core of the paper to a level of detail that is necessary to appreciate the results and make sense of them.
        \item The full details can be provided either with the code, in appendix, or as supplemental material.
    \end{itemize}

\item {\bf Experiment statistical significance}
    \item[] Question: Does the paper report error bars suitably and correctly defined or other appropriate information about the statistical significance of the experiments?
    \item[] Answer: \answerYes{} %
    \item[] Justification: All reported experiments are averaged over three random seeds, covering stochasticity from LLM sampling, GP initialization, and MCTS rollouts, to ensure that the reported results reflect stable performance rather than single-run fluctuations.
    \item[] Guidelines:
    \begin{itemize}
        \item The answer \answerNA{} means that the paper does not include experiments.
        \item The authors should answer \answerYes{} if the results are accompanied by error bars, confidence intervals, or statistical significance tests, at least for the experiments that support the main claims of the paper.
        \item The factors of variability that the error bars are capturing should be clearly stated (for example, train/test split, initialization, random drawing of some parameter, or overall run with given experimental conditions).
        \item The method for calculating the error bars should be explained (closed form formula, call to a library function, bootstrap, etc.)
        \item The assumptions made should be given (e.g., Normally distributed errors).
        \item It should be clear whether the error bar is the standard deviation or the standard error of the mean.
        \item It is OK to report 1-sigma error bars, but one should state it. The authors should preferably report a 2-sigma error bar than state that they have a 96\% CI, if the hypothesis of Normality of errors is not verified.
        \item For asymmetric distributions, the authors should be careful not to show in tables or figures symmetric error bars that would yield results that are out of range (e.g., negative error rates).
        \item If error bars are reported in tables or plots, the authors should explain in the text how they were calculated and reference the corresponding figures or tables in the text.
    \end{itemize}

\item {\bf Experiments compute resources}
    \item[] Question: For each experiment, does the paper provide sufficient information on the computer resources (type of compute workers, memory, time of execution) needed to reproduce the experiments?
    \item[] Answer: \answerYes{} %
    \item[] Justification: The hardware configuration (GPUs for vLLM-based LLM serving and CPU for non-LLM solvers) is reported in Appendix~\ref{app:compute}. Per-task average search time for every (solver, domain) configuration is reported in the Time column of Table~\ref{tab:main-results}.
    \item[] Guidelines:
    \begin{itemize}
        \item The answer \answerNA{} means that the paper does not include experiments.
        \item The paper should indicate the type of compute workers CPU or GPU, internal cluster, or cloud provider, including relevant memory and storage.
        \item The paper should provide the amount of compute required for each of the individual experimental runs as well as estimate the total compute. 
        \item The paper should disclose whether the full research project required more compute than the experiments reported in the paper (e.g., preliminary or failed experiments that didn't make it into the paper). 
    \end{itemize}
    
\item {\bf Code of ethics}
    \item[] Question: Does the research conducted in the paper conform, in every respect, with the NeurIPS Code of Ethics \url{https://neurips.cc/public/EthicsGuidelines}?
    \item[] Answer: \answerYes{} %
    \item[] Justification: The authors have reviewed the NeurIPS Code of Ethics. The work uses publicly released benchmarks (LLM-SRBench) and open-weight LLMs.
    \item[] Guidelines:
    \begin{itemize}
        \item The answer \answerNA{} means that the authors have not reviewed the NeurIPS Code of Ethics.
        \item If the authors answer \answerNo, they should explain the special circumstances that require a deviation from the Code of Ethics.
        \item The authors should make sure to preserve anonymity (e.g., if there is a special consideration due to laws or regulations in their jurisdiction).
    \end{itemize}

\item {\bf Broader impacts}
    \item[] Question: Does the paper discuss both potential positive societal impacts and negative societal impacts of the work performed?
    \item[] Answer: \answerYes{} %
    \item[] Justification: Section~\ref{sec:intro} motivates DASS for scientific discovery and interpretable modeling, and Section~\ref{4exp:budget} shows it lowers the search cost of symbolic regression. As a search-guidance module operating on numerical scientific data, it has no foreseeable direct negative societal impact.
    \item[] Guidelines:
    \begin{itemize}
        \item The answer \answerNA{} means that there is no societal impact of the work performed.
        \item If the authors answer \answerNA{} or \answerNo, they should explain why their work has no societal impact or why the paper does not address societal impact.
        \item Examples of negative societal impacts include potential malicious or unintended uses (e.g., disinformation, generating fake profiles, surveillance), fairness considerations (e.g., deployment of technologies that could make decisions that unfairly impact specific groups), privacy considerations, and security considerations.
        \item The conference expects that many papers will be foundational research and not tied to particular applications, let alone deployments. However, if there is a direct path to any negative applications, the authors should point it out. For example, it is legitimate to point out that an improvement in the quality of generative models could be used to generate Deepfakes for disinformation. On the other hand, it is not needed to point out that a generic algorithm for optimizing neural networks could enable people to train models that generate Deepfakes faster.
        \item The authors should consider possible harms that could arise when the technology is being used as intended and functioning correctly, harms that could arise when the technology is being used as intended but gives incorrect results, and harms following from (intentional or unintentional) misuse of the technology.
        \item If there are negative societal impacts, the authors could also discuss possible mitigation strategies (e.g., gated release of models, providing defenses in addition to attacks, mechanisms for monitoring misuse, mechanisms to monitor how a system learns from feedback over time, improving the efficiency and accessibility of ML).
    \end{itemize}
    
\item {\bf Safeguards}
    \item[] Question: Does the paper describe safeguards that have been put in place for responsible release of data or models that have a high risk for misuse (e.g., pre-trained language models, image generators, or scraped datasets)?
    \item[] Answer: \answerNA{} %
    \item[] Justification: The paper proposes an algorithmic framework only. ; it does not introduce pretrained models, generative systems, or scraped datasets, and therefore poses no elevated misuse risk.
    \item[] Guidelines:
    \begin{itemize}
        \item The answer \answerNA{} means that the paper poses no such risks.
        \item Released models that have a high risk for misuse or dual-use should be released with necessary safeguards to allow for controlled use of the model, for example by requiring that users adhere to usage guidelines or restrictions to access the model or implementing safety filters. 
        \item Datasets that have been scraped from the Internet could pose safety risks. The authors should describe how they avoided releasing unsafe images.
        \item We recognize that providing effective safeguards is challenging, and many papers do not require this, but we encourage authors to take this into account and make a best faith effort.
    \end{itemize}

\item {\bf Licenses for existing assets}
    \item[] Question: Are the creators or original owners of assets (e.g., code, data, models), used in the paper, properly credited and are the license and terms of use explicitly mentioned and properly respected?
    \item[] Answer: \answerYes{} %
    \item[] Justification: All used assets are credited via citations in the main text and Appendix~\ref{app:base-solvers}; the code packages are released under permissive open-source licenses (Apache~2.0, BSD, MIT) and the LLMs follow their model-card terms (Qwen License, Gemma Terms of Use, OpenAI Terms of Use). All assets are used non-commercially for academic research.
    \item[] Guidelines:
    \begin{itemize}
        \item The answer \answerNA{} means that the paper does not use existing assets.
        \item The authors should cite the original paper that produced the code package or dataset.
        \item The authors should state which version of the asset is used and, if possible, include a URL.
        \item The name of the license (e.g., CC-BY 4.0) should be included for each asset.
        \item For scraped data from a particular source (e.g., website), the copyright and terms of service of that source should be provided.
        \item If assets are released, the license, copyright information, and terms of use in the package should be provided. For popular datasets, \url{paperswithcode.com/datasets} has curated licenses for some datasets. Their licensing guide can help determine the license of a dataset.
        \item For existing datasets that are re-packaged, both the original license and the license of the derived asset (if it has changed) should be provided.
        \item If this information is not available online, the authors are encouraged to reach out to the asset's creators.
    \end{itemize}

\item {\bf New assets}
    \item[] Question: Are new assets introduced in the paper well documented and is the documentation provided alongside the assets?
    \item[] Answer: \answerNA{} %
    \item[] Justification: All experiments are conducted on the public LSR-Synth benchmark from LLM-SRBench~\citep{llmsrbench}; the DASS implementation will be released upon acceptance as noted in Q5.
    \item[] Guidelines:
    \begin{itemize}
        \item The answer \answerNA{} means that the paper does not release new assets.
        \item Researchers should communicate the details of the dataset\slash code\slash model as part of their submissions via structured templates. This includes details about training, license, limitations, etc. 
        \item The paper should discuss whether and how consent was obtained from people whose asset is used.
        \item At submission time, remember to anonymize your assets (if applicable). You can either create an anonymized URL or include an anonymized zip file.
    \end{itemize}

\item {\bf Crowdsourcing and research with human subjects}
    \item[] Question: For crowdsourcing experiments and research with human subjects, does the paper include the full text of instructions given to participants and screenshots, if applicable, as well as details about compensation (if any)? 
    \item[] Answer: \answerNA{} %
    \item[] Justification: The paper does not involve crowdsourcing or research with human subjects.
    \item[] Guidelines:
    \begin{itemize}
        \item The answer \answerNA{} means that the paper does not involve crowdsourcing nor research with human subjects.
        \item Including this information in the supplemental material is fine, but if the main contribution of the paper involves human subjects, then as much detail as possible should be included in the main paper. 
        \item According to the NeurIPS Code of Ethics, workers involved in data collection, curation, or other labor should be paid at least the minimum wage in the country of the data collector. 
    \end{itemize}

\item {\bf Institutional review board (IRB) approvals or equivalent for research with human subjects}
    \item[] Question: Does the paper describe potential risks incurred by study participants, whether such risks were disclosed to the subjects, and whether Institutional Review Board (IRB) approvals (or an equivalent approval/review based on the requirements of your country or institution) were obtained?
    \item[] Answer: \answerNA{} %
    \item[] Justification: No human subjects are involved.
    \item[] Guidelines:
    \begin{itemize}
        \item The answer \answerNA{} means that the paper does not involve crowdsourcing nor research with human subjects.
        \item Depending on the country in which research is conducted, IRB approval (or equivalent) may be required for any human subjects research. If you obtained IRB approval, you should clearly state this in the paper. 
        \item We recognize that the procedures for this may vary significantly between institutions and locations, and we expect authors to adhere to the NeurIPS Code of Ethics and the guidelines for their institution. 
        \item For initial submissions, do not include any information that would break anonymity (if applicable), such as the institution conducting the review.
    \end{itemize}

\item {\bf Declaration of LLM usage}
    \item[] Question: Does the paper describe the usage of LLMs if it is an important, original, or non-standard component of the core methods in this research? Note that if the LLM is used only for writing, editing, or formatting purposes and does \emph{not} impact the core methodology, scientific rigor, or originality of the research, declaration is not required.
    \item[] Answer: \answerYes{} %
    \item[] Justification: GPT-4o is used to initialize the basis term set in Section~\ref{3method:InitSubspace} (prompt in Appendix~\ref{app:prompt}); Qwen3.5-9B and Gemma-4-31B serve as the LLM backends for LLM-SR and LaSR in Appendix~\ref{app:llm-backbones}.
    \item[] Guidelines:
    \begin{itemize}
        \item The answer \answerNA{} means that the core method development in this research does not involve LLMs as any important, original, or non-standard components.
        \item Please refer to our LLM policy in the NeurIPS handbook for what should or should not be described.
    \end{itemize}

\end{enumerate}

\end{document}